%% file: main.tex
\documentclass[letterpaper,twocolumn,10pt]{article}
\usepackage{usenix2019_v3}

\usepackage{caption}
\usepackage{color}
\usepackage{multirow}
\usepackage{url}
\usepackage{rotating}
\usepackage{authblk}
\usepackage{array}
\usepackage{algorithmicx,algorithm}
\usepackage[noend]{algpseudocode}
\usepackage{subfigure}
\usepackage{enumitem} 
\usepackage{times}
\usepackage{xspace}
\usepackage{latexsym}
\usepackage{amsmath}
\usepackage{booktabs}
\usepackage{tabularx}
\usepackage{balance}
\usepackage{graphicx}
\usepackage{svg}
\usepackage{pifont}
\svgsetup{
    inkscapepath=i/svg-inkscape/
}
\svgpath{{svg/}}
\usepackage{amssymb}
\usepackage{etoolbox}
\usepackage[all]{nowidow}
\usepackage{colortbl}
\usepackage[normalem]{ulem}

\setlist{nolistsep}
\newcommand{\parabf}[1]{\medskip\noindent\textbf{#1}}
\newcommand{\parait}[1]{\medskip\noindent\textit{#1}}
\newcommand{\paraf}[1]{\noindent\textbf{#1}}
\newcommand{\cut}[1]{}

\newcommand{\sysname}{ReLibra\xspace}

\definecolor{applegreen}{rgb}{0.55, 0.71, 0.0}
\definecolor{forestgreen}{rgb}{0.13, 0.55, 0.13}
\definecolor{antiquewhite}{rgb}{0.98, 0.92, 0.84}
\newlength{\titleblockheight}
\makeatletter
\patchcmd{\@maketitle}
  {\vbox to 1.3in}
  {\vbox to \titleblockheight}
  {}{}

\begin{document}
\sloppy
\date{}

\title{\sysname: Routing-Replay-Guided Load Balancing for MoE Training\\in Reinforcement Learning}

\author{
    Chao Jin$^{\ast}$\qquad Xinming Wei$^{\ast}$\qquad Yinmin Zhong$^{\ast}$\qquad Chengxu Yang$^{\dagger}$\qquad Bingyang Wu$^{\ast}$ \authorcr
    Ruidong Zhu$^{\ast}$\qquad Zili Zhang$^{\ast}$\qquad Yuliang Liu$^{\dagger}$\qquad Xin Jin$^{\ast}$ \authorcr
    $^{\ast}$\textit{School of Computer Science, Peking University} \qquad $^{\dagger}$\textit{Independent Researcher}
}

\maketitle
\pagestyle{plain}

\captionsetup[figure]{font=small}
\captionsetup[table]{font=small}

\input{sections/abstract}
\input{sections/introduction}
\input{sections/background}
\input{sections/overview}
\input{sections/design}
\input{sections/implementation}
\input{sections/evaluation}
\input{sections/discussion}
\input{sections/related}

\input{sections/conclusion}

\label{lastpage}

{
\bibliographystyle{ieeetr}
\bibliography{reference}}


\end{document}

%% file: sections/abstract.tex
\begin{abstract}

Load imbalance is a long-standing challenge in Mixture-of-Experts (MoE) training and is exacerbated in 
reinforcement learning (RL) for LLMs, where hot experts can 
shift frequently across micro-batches. Existing MoE training systems rely on historical loads to predict 
future expert demand, making them less effective under sharp fluctuations. 

We propose \sysname, an MoE RL training system that exploits a unique opportunity in RL's rollout-training workflow, 
\uline{routing replay}, to enable fine-grained load balancing at micro-batch granularity. 
Because rollout and training process the same tokens with the same MoE parameters, 
the token-to-expert routing decisions are known before training starts. 
Leveraging this information, \sysname places two MoE load-balancing mechanisms 
at inter- and intra-batch timescales, matching their communication patterns to hierarchical network bandwidths. 
At the inter-batch timescale, \sysname performs \uline{expert reordering} to redistribute experts for batch-level 
cross-node balancing; at the intra-batch timescale, it dynamically performs 
\uline{expert replication} within a node to absorb micro-batch-level load fluctuations. 
Experiments on diverse MoE LLMs and RL workloads show that \sysname improves training throughput 
by up to 1.6$\times$ over Megatron-LM and by up to 1.2$\times$ over EPLB, even when EPLB is given oracle loads. 
Moreover, \sysname remains within 6\%--10\% of the throughput of an idealized balanced baseline.

\end{abstract}

%% file: sections/introduction.tex
\section{Introduction}
\label{sec:introduction}

Mixture-of-Experts (MoE)~\cite{shazeer2017outrageously, lepikhin2021gshard, fedus2022switch} has emerged as a
mainstream architecture for scaling large language models (LLMs)~\cite{dbrx, jiang2024mixtral, liu2024deepseek, yang2025qwen3, team2025kimi, chen2025minimax},
as it increases model capacity without proportionally increasing computation by sparsely routing tokens to a small set 
of \uline{experts}~\cite{du2022glam, rajbhandari2022deepspeed}. However, MoE training in both pretraining and 
RL post-training~\cite{ouyang2022training, guo2025deepseek} suffers from 
load imbalance~\cite{nie2023flexmoe, zhai2023smartmoe, zhao2025micromoe, zhang2025popfetcher, liu2026laer}, 
because dynamic routing can overload GPUs hosting hot experts under expert parallelism (EP)~\cite{liu2024deepseek, rajbhandari2022deepspeed, liu2025moe}, 
creating stragglers and reducing training efficiency.

While MoE pretraining typically exhibits a relatively stable imbalance pattern, 
MoE RL training shows a more fluctuating one (\S~\ref{sec:background:characterization}). 
We use the term imbalance pattern to describe how expert load distribution varies across micro-batches, 
where a training batch is split into multiple micro-batches that are processed separately to reduce activation memory. 
In RL training, the set of hot experts can change substantially from one micro-batch to the next, 
whereas in pretraining it remains relatively stable. 
This difference arises from two main reasons. 
First, while MoE pretraining is conducted on web-scale, diverse corpora, MoE RL training typically
targets narrower domains~\cite{guo2025deepseek, liu2025deepseek-v3-2, team2025kimi} (e.g., 
math~\cite{yu2025dapo}, coding~\cite{li2022competition}, and instruction following~\cite{pyatkin2025generalizing}).
As a result, micro-batches from different domains in RL training tend to activate different sets of hot experts.
Second, MoE pretraining employs algorithmic load-balancing mechanisms to prevent some experts from being undertrained, 
such as auxiliary balancing losses~\cite{shazeer2017outrageously, lepikhin2021gshard, fedus2022switch} and 
routing bias~\cite{liu2024deepseek}, whereas these mechanisms are removed during RL 
training~\cite{shao2024deepseekmath, yu2025dapo, chen2025minimax}, since experts have already been shaped during pretraining. 
Without such mechanisms, expert load distributions in RL training become more skewed and unstable, even within a single domain.

\begin{figure}[t!]
    \centering
    \includegraphics[width=\linewidth]{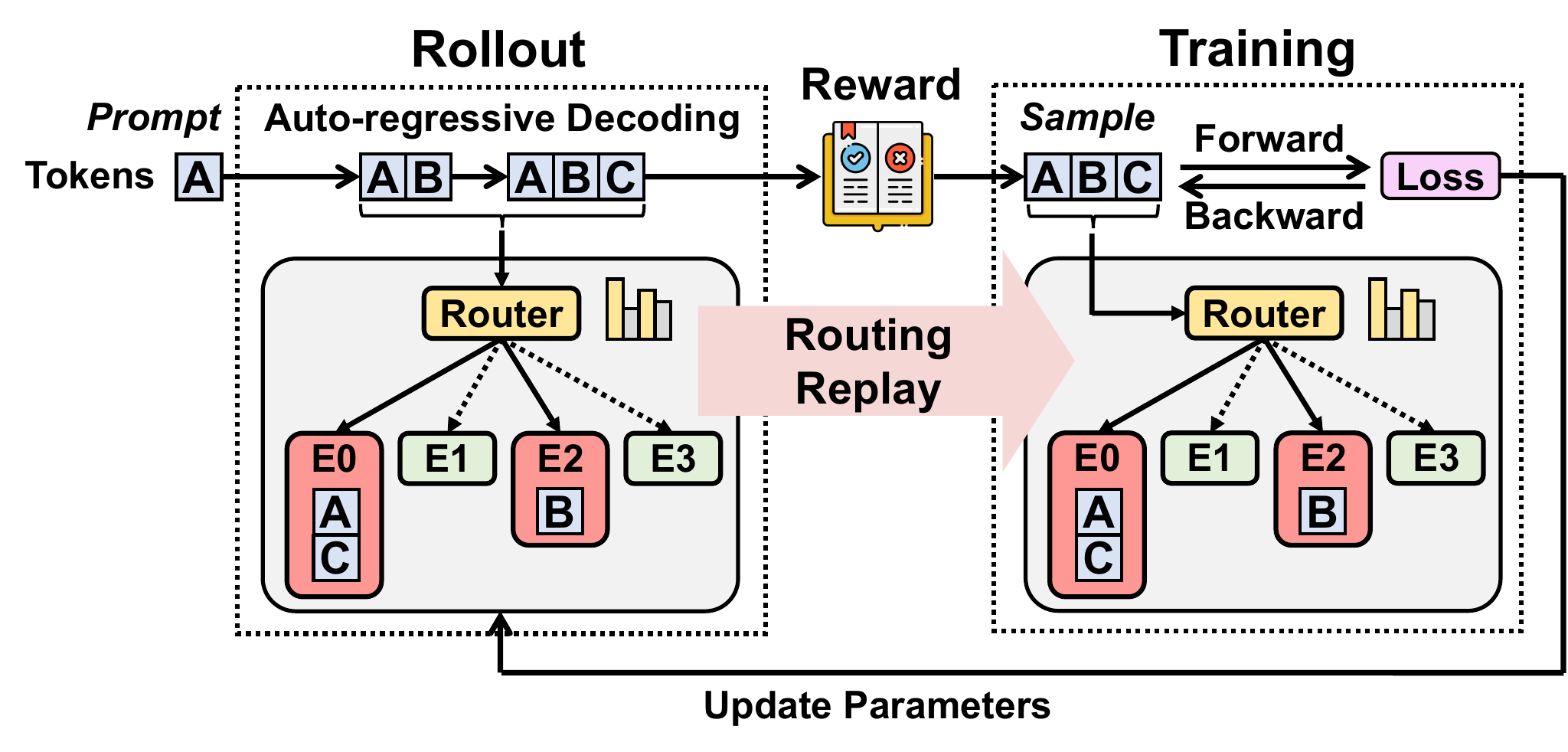}
    \vspace*{-0.2in}
    \caption{MoE RL training workflow and the opportunity for routing replay.}
    \vspace*{-0.1in}
    \label{fig:intro:routing_replay}
\end{figure}

Existing MoE training systems struggle to cope with such fluctuating imbalance patterns because they are primarily designed for pretraining, 
where expert load distributions are relatively stable and therefore predictable.
For example, EPLB~\cite{eplb} periodically replicates hot experts based on recent load statistics, 
typically once every one or several training batches to amortize the reconfiguration overhead.  
PopFetcher~\cite{zhang2025popfetcher} prefetches experts using activation patterns from recent training batches in adjacent layers. 
Despite their different mechanisms, both systems rely on the same assumption: future expert demand can be predicted from 
recent history. When expert load distributions fluctuate sharply at micro-batch granularity in RL training, 
this assumption breaks down, limiting the effectiveness of prior solutions.

Although RL makes expert load distributions more unstable, its workflow also creates a unique opportunity for fine-grained load balancing via \uline{routing replay}.
As shown in Figure~\ref{fig:intro:routing_replay}, RL consists of a rollout stage that generates responses and a training stage that updates the model on the generated samples.
Since both stages process the same token sequence with the same MoE parameters, they produce the same routing scores and, theoretically,
the same top-$k$ experts. Thus, the exact token-to-expert routing decisions are determined during rollout and can be replayed in training.
Moreover, while some RL algorithms~\cite{mnih2015human, haarnoja2018soft} perform multiple updates per rollout, recent industrial
practice~\cite{ma2025stabilizing, liu2025deepseek-v3-2, zheng2025stabilizing} has also adopted routing replay so that experts are
updated with the exact tokens they generated, which may improve algorithmic training stability.

\begin{figure}[t!]
    \centering
    \includegraphics[width=0.9\linewidth]{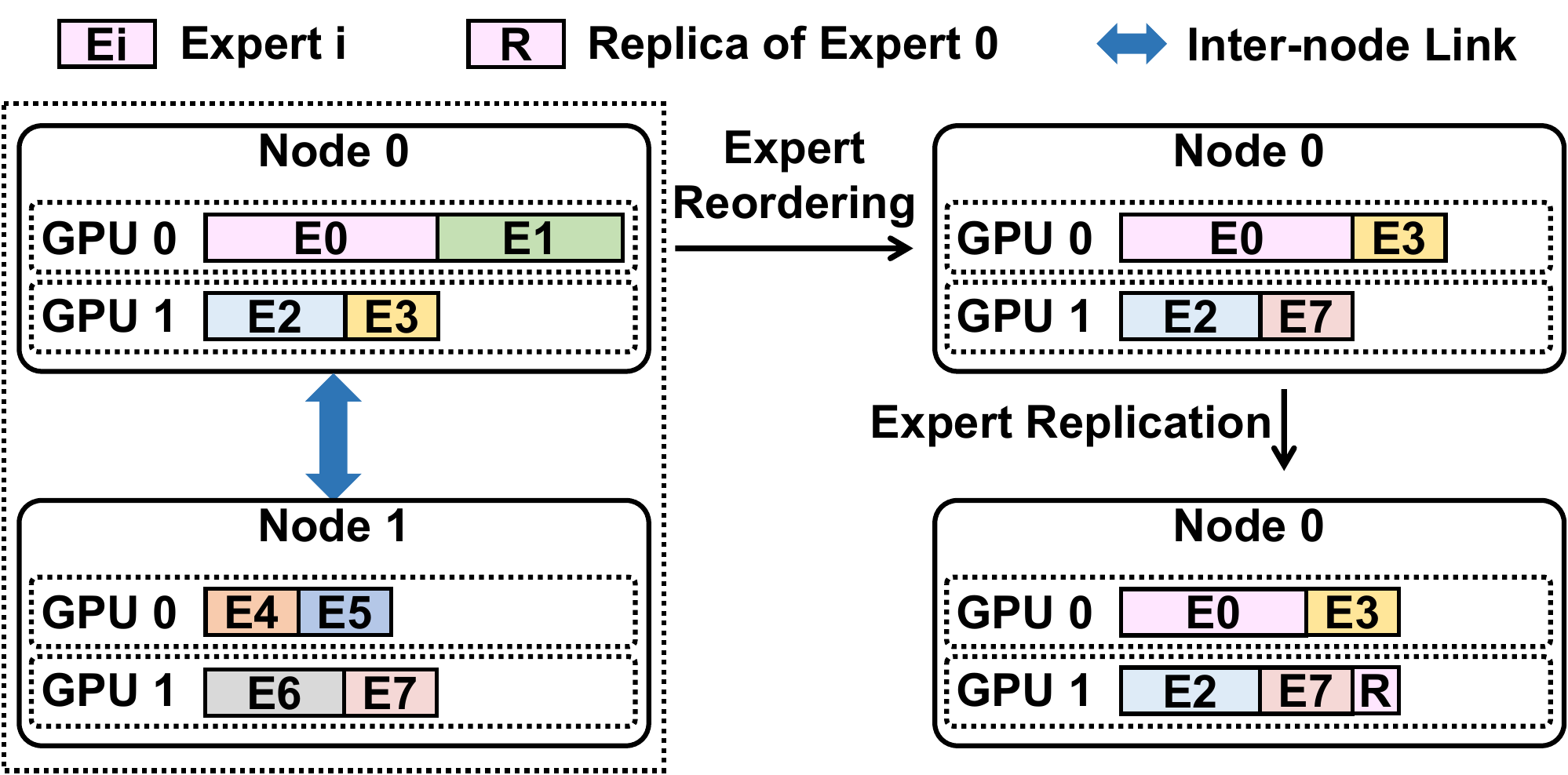}
    \vspace*{-0.1in}
    \caption{Load-balancing mechanisms for MoE training.}
    \vspace*{-0.1in}
    \label{fig:intro:balancing_space}
\end{figure}

However, even with routing known in advance, finding the optimal load-balancing plan for each micro-batch and MoE layer 
remains difficult. As shown in Figure~\ref{fig:intro:balancing_space}, MoE training systems provide two main mechanisms: 
\uline{expert reordering}, which redistributes experts across the EP group, and \uline{expert replication}, which adds 
replicas for overloaded experts. Applying them at micro-batch granularity results in a large combinatorial optimization problem, 
and even simplified variants are NP-hard~\cite{graham1979optimization}, making exact solving impractical. 

To this end, we propose \sysname, an MoE RL training system that exploits the routing replay opportunity in RL to 
address fluctuating load imbalance at micro-batch granularity. 
The key insight of \sysname is to place the two 
system-level load-balancing mechanisms at inter- and intra-batch timescales by matching their communication 
patterns with the hierarchical network bandwidths in training clusters.

To scale MoE models with massive numbers of experts, cross-node expert parallelism is
widely adopted~\cite{liu2024deepseek, team2025kimi, liu2025moe, chen2025minimax}. However, because imbalance
patterns vary rapidly across micro-batches, frequent expert reordering may incur expensive communication over
low-bandwidth inter-node links. By contrast, expert replication can often be restricted within a node,
making it more efficient to adjust at micro-batch granularity. This asymmetry motivates
a hierarchical design: \sysname applies inter-batch (per training batch) expert reordering for coarse-grained 
cross-node balancing and intra-batch (per micro-batch) expert replication for fine-grained local adaptation.
This decomposition makes the original problem tractable, which better matches the network topology
and achieves near-optimal performance in practice across diverse MoE RL workloads, as shown in our evaluation (\S~\ref{sec:evaluation:overall}).

At the inter-batch timescale, \sysname jointly optimizes computation and communication imbalance using routing information aggregated over the training batch. 
It employs a swap-based simulated annealing algorithm to efficiently search for an expert reordering plan that minimizes MoE execution time across the EP group, 
followed by a second round of optimization to improve data locality.

At the intra-batch timescale, \sysname dynamically replicates experts within a node to absorb micro-batch-level load 
fluctuations. Its key design is a layer-shared replica buffer that stores replicas for only one layer 
at a time and is reused across all layers, incurring negligible memory overhead. 
Replica synchronization is overlapped with attention computation to avoid affecting the critical path. On top of this design, 
\sysname optimizes replica placement and token splitting across replicas via a mixed-integer linear program (MILP).

\begin{figure}[t!]
    \centering
    \includegraphics[width=0.9\linewidth]{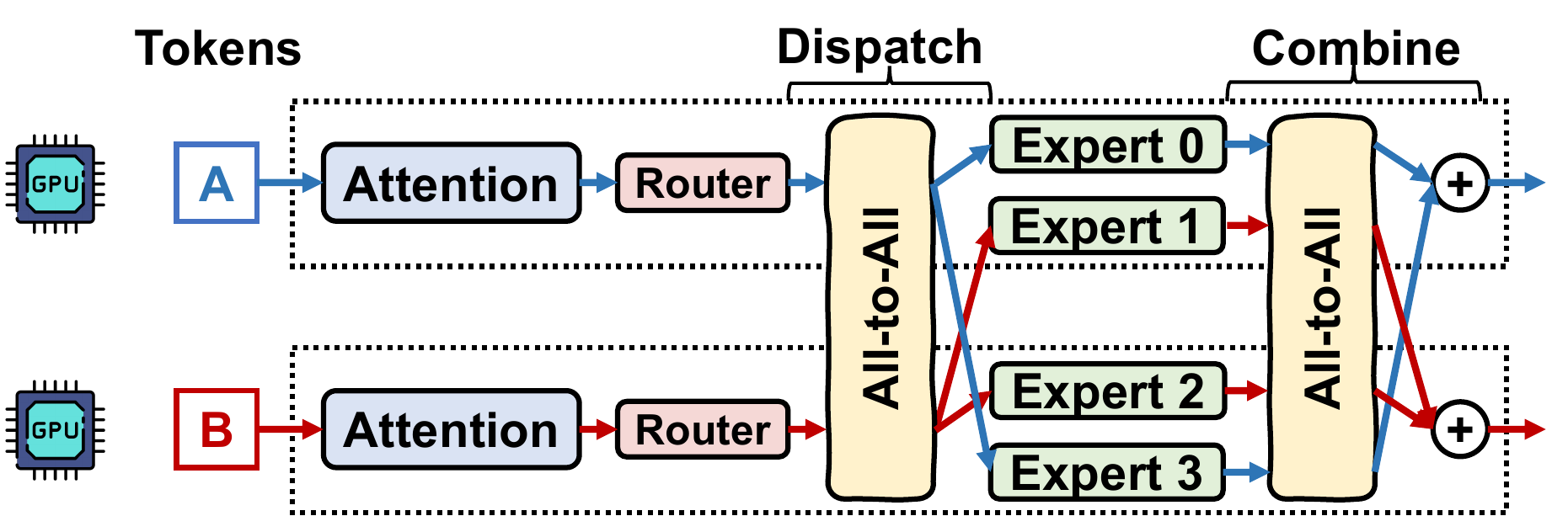}
    \vspace*{-0.1in}
    \caption{MoE layer and expert parallelism.}
    \vspace*{-0.1in}
    \label{fig:background:moe_ep}
\end{figure}

In summary, we make the following contributions. 
\begin{itemize}[leftmargin=*]
    \item We propose \sysname that exploits the routing replay opportunity in MoE RL training to 
    address fluctuating load imbalance at micro-batch granularity. 
    \item We place expert reordering and replication at inter- and intra-batch timescales, respectively, 
    to match their communication patterns to the network hierarchy of training clusters, 
    thereby decomposing the original problem into two tractable sub-problems. Built on efficient 
    algorithms and lightweight system designs, \sysname achieves fine-grained load balancing with 
    negligible overhead.
    \item We implement \sysname and evaluate it on MoE LLMs of varying scales across diverse RL domains. 
    Experimental results show that \sysname improves training throughput by up to 1.6$\times$ over Megatron-LM, 
    a state-of-the-art MoE training system, and by up to 1.2$\times$ over EPLB, even when EPLB is provided with 
    oracle loads. Moreover, \sysname attains 90\%--94\% of the throughput of an idealized balanced baseline.
\end{itemize}


%% file: sections/background.tex
\section{Background and Motivation}
\label{sec:background}

\subsection{Background}
\label{sec:background:background}

\paraf{MoE training.} MoE extends the Transformer~\cite{vaswani2017attention} architecture by replacing the dense 
feed-forward network (FFN) with a sparse MoE layer containing multiple FFN experts and a trainable router. 
For each input token, the router selects a small set of experts (i.e., top-$k$ experts), 
and the outputs of the selected experts are combined to produce the final result. 
This sparse activation pattern substantially increases model capacity without proportionally 
increasing per-token computation, making MoE a practical approach for 
scaling LLMs~\cite{du2022glam, rajbhandari2022deepspeed, jiang2024mixtral, liu2024deepseek, yang2025qwen3, team2025kimi, chen2025minimax}. 

The large parameter count of MoE models introduces significant memory pressure during training, 
making distributed training necessary. Expert parallelism (EP), tailored for MoE models, 
distributes different experts of an MoE layer across GPUs and relies on all-to-all communication to 
dispatch tokens and combine their outputs as shown in Figure~\ref{fig:background:moe_ep}. 
Recent MoE models continue to scale parameter count by increasing the number of experts. 
For example, Qwen3-235B-A22B~\cite{yang2025qwen3} uses 128 experts, while DeepSeek-V3-671B~\cite{liu2024deepseek} uses 256 experts. 
Consequently, cross-node EP is increasingly common in practice. 
Existing MoE training frameworks also combine EP with conventional parallelism strategies, 
such as data parallelism (DP)~\cite{rajbhandari2020zero}, tensor parallelism (TP)~\cite{megatron-lm-arxiv}, 
and pipeline parallelism (PP)~\cite{huang2019gpipe, narayanan2019pipedream}, 
to jointly meet the scalability and memory requirements of large-scale training. 

\begin{table}[t!]
    \centering
    \arrayrulewidth=0.5pt
    \extrarowheight=1pt
    \resizebox{0.85\linewidth}{!} {
        \begin{tabular}{@{}cc@{}}
            \arrayrulecolor{black}\hline
            \arrayrulecolor{black}\hline
            Domain                & Dataset \\ \hline
            Reasoning             & DAPO-Math-17k~\cite{yu2025dapo}, GPQA~\cite{rein2024gpqa} \\
            Instruction Following & IFBench~\cite{pyatkin2025generalizing} \\
            Coding                & CodeForces~\cite{li2022competition} \\ 
            Mixed                 & Mixture of the above three domains \\ 
                                  & as suggested by DeepSeek-V3.2~\cite{liu2025deepseek-v3-2} \\
            \arrayrulecolor{black}\hline
            \arrayrulecolor{black}\hline
        \end{tabular}
    }
    \vspace{-0.1in}
    \caption{Domains and datasets used in our experiments.}
    \vspace{-0.1in}
    \label{tab:background:datasets}
\end{table}

\begin{figure}[t!]
    \centering
    \includegraphics[width=0.85\linewidth]{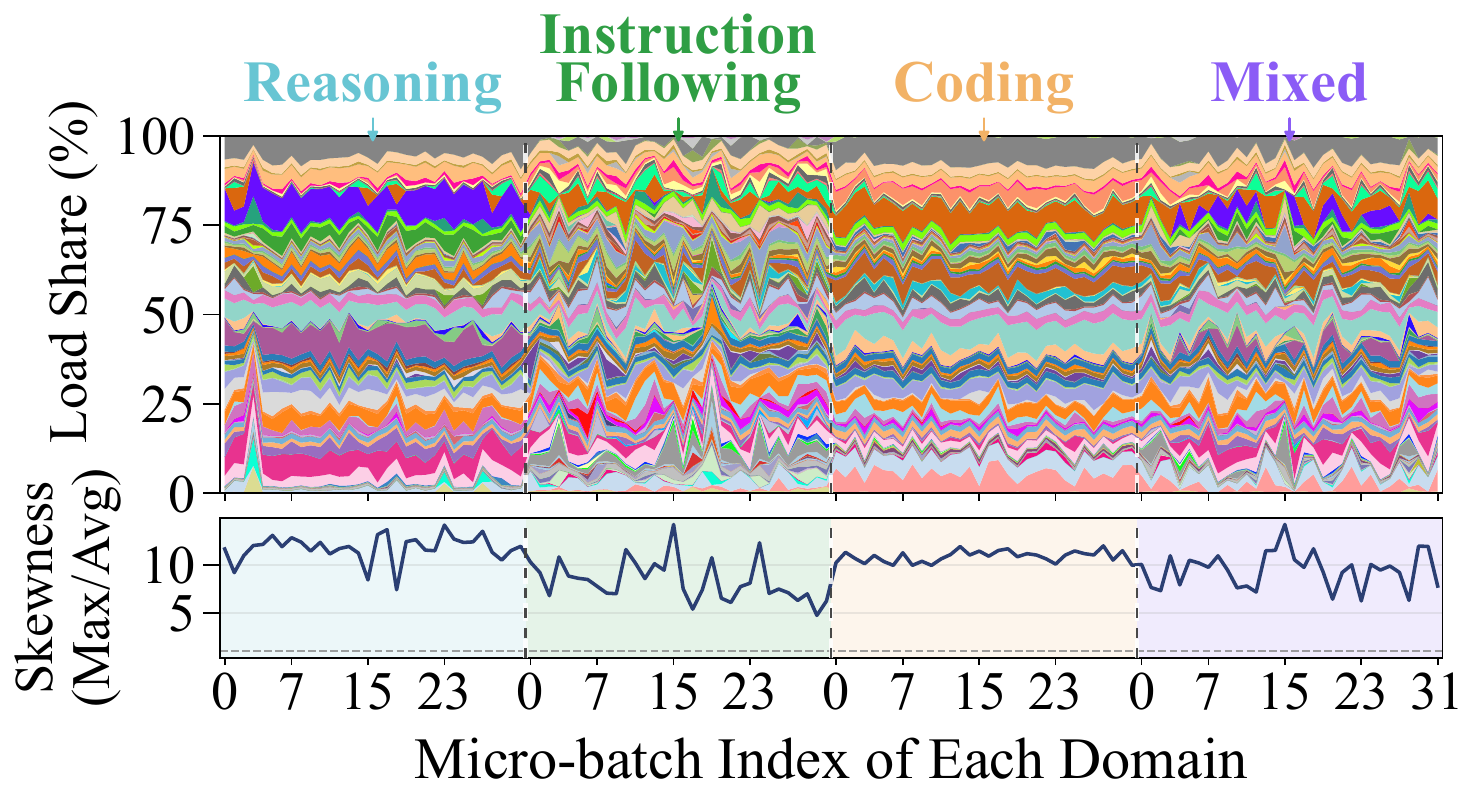}
    \vspace*{-0.1in}
    \caption{Dynamics of expert loads for one MoE layer across micro-batches from different domains.}
    \vspace*{-0.1in}
    \label{fig:background:load_dynamics}
\end{figure}

\parabf{RL training workflow.} RL for LLMs has become a key post-training technique for enhancing domain-specific capabilities by optimizing the model 
with external feedback on its self-generated responses, such as human feedback~\cite{ouyang2022training}, 
rule-based verifiable rewards~\cite{guo2025deepseek, shao2024deepseekmath}, 
and generative reward models~\cite{liu2025deepseek-v3-2, team2025kimi}. 
Such capabilities include alignment with human preferences~\cite{ouyang2022training} and improved reasoning ability~\cite{guo2025deepseek}. 

Each RL iteration can be viewed as a two-stage loop, as shown in Figure~\ref{fig:intro:routing_replay}: rollout and training. 
In the rollout stage, the model takes a set of prompts and generates one or more responses per prompt via auto-regressive 
decoding~\cite{yu2022orca, wu2023fast}. Post-rollout processing scores each response with a reward using either rule-based functions or a learned reward model. 
Some RL algorithms, such as PPO~\cite{schulman2017proximal}, also use a critic model to estimate fine-grained token-level rewards. 
Others like GRPO~\cite{shao2024deepseekmath} use approximation strategies to reduce computational costs. A forward pass of 
a reference model may also be performed to compute the Kullback--Leibler (KL) divergence from the model's initial state to stabilize training. 
The prompts and responses are then concatenated into training samples, which, together with their reward signals, 
form a training batch. In the training stage, this batch is divided into micro-batches, each of which performs forward and 
backward passes and accumulates gradients until the full batch is processed. The accumulated gradients 
are then used to update the model parameters, which are synchronized back to the rollout workers for the next iteration.

While recent work targets the rollout stage~\cite{hu2026taming, qin2025seer}, our work focuses on a complementary 
bottleneck: improving the efficiency of the training stage itself, specifically the fluctuating load imbalance problem for MoE models. 

\subsection{Load Imbalance Characterization}
\label{sec:background:characterization}

\begin{figure}[t!]
    \centering
    \includegraphics[width=0.95\linewidth]{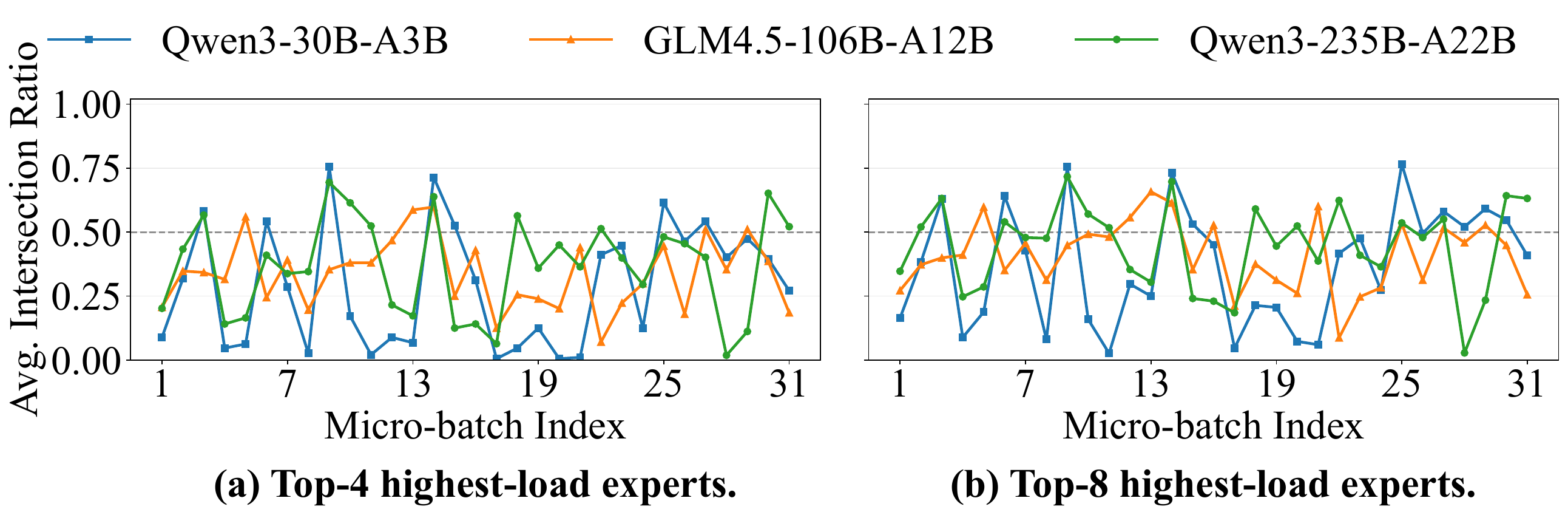}
    \vspace*{-0.1in}
    \caption{Average intersection ratio of top-$k$ highest-load experts between adjacent micro-batches. 
    Lower values indicate more diverse hot experts and therefore more fluctuating load imbalance.}
    \vspace*{-0.1in}
    \label{fig:background:intersection_ratio}
\end{figure}

We characterize the imbalance patterns in MoE RL training using three representative workloads, 
along with a mixed workload constructed from them following DeepSeek-V3.2~\cite{liu2025deepseek-v3-2}. 
Table~\ref{tab:background:datasets} summarizes the domains and datasets used in our experiments. 
During rollout, we set the maximum response length to 8K tokens and 
generate four samples per prompt using GRPO~\cite{shao2024deepseekmath}, 
with temperature 1.0 and top-$p$ 1.0. 
For all domains, we use an EP size of 32, a global batch size of 1024, and 32 micro-batches.

Figure~\ref{fig:background:load_dynamics} illustrates the imbalance pattern dynamics, 
including per-expert load share and skewness, across consecutive micro-batches of 
Qwen3-235B-A22B~\cite{yang2025qwen3}. Across different domains, the same color denotes the same expert. 
We define the load skewness of a micro-batch as the ratio between the maximum expert load and 
the average expert load over all experts in that micro-batch. 
Figure~\ref{fig:background:load_dynamics} reveals fluctuating load imbalance in two complementary ways. 
First, the load share of an individual expert varies substantially from one micro-batch to the 
next, indicating that the hot experts are not stable even within the same 
domain. Second, the skewness of the expert load distribution is dynamic. 
Different domains exhibit different skewness dynamics and activate different subsets of experts. 
When several domains are mixed for RL training, hot experts can shift rapidly across micro-batches.

To quantify the temporal fluctuation of load imbalance, we measure the intersection ratio of the top-$k$ 
highest-load experts between adjacent micro-batches, defined as $\frac{|H_t^k \cap H_{t+1}^k|}{k}$, 
where $H_t^k$ denotes the set of top-$k$ highest-load experts in micro-batch $t$. 
A lower intersection ratio indicates more drastic changes in hot experts across micro-batches.
Figure~\ref{fig:background:intersection_ratio} reports the intersection 
ratio of the top-$4$ and top-$8$ highest-load experts on the mixed domain, with 
results averaged over all MoE layers for each model. Across three MoE models ranging from 30B to 235B parameters, 
the intersection ratio stays below 0.5 for most cases, indicating that more than half of the hot experts change 
from one micro-batch to the next. 

\begin{figure}[t!]
    \centering
    \includegraphics[width=0.7\linewidth]{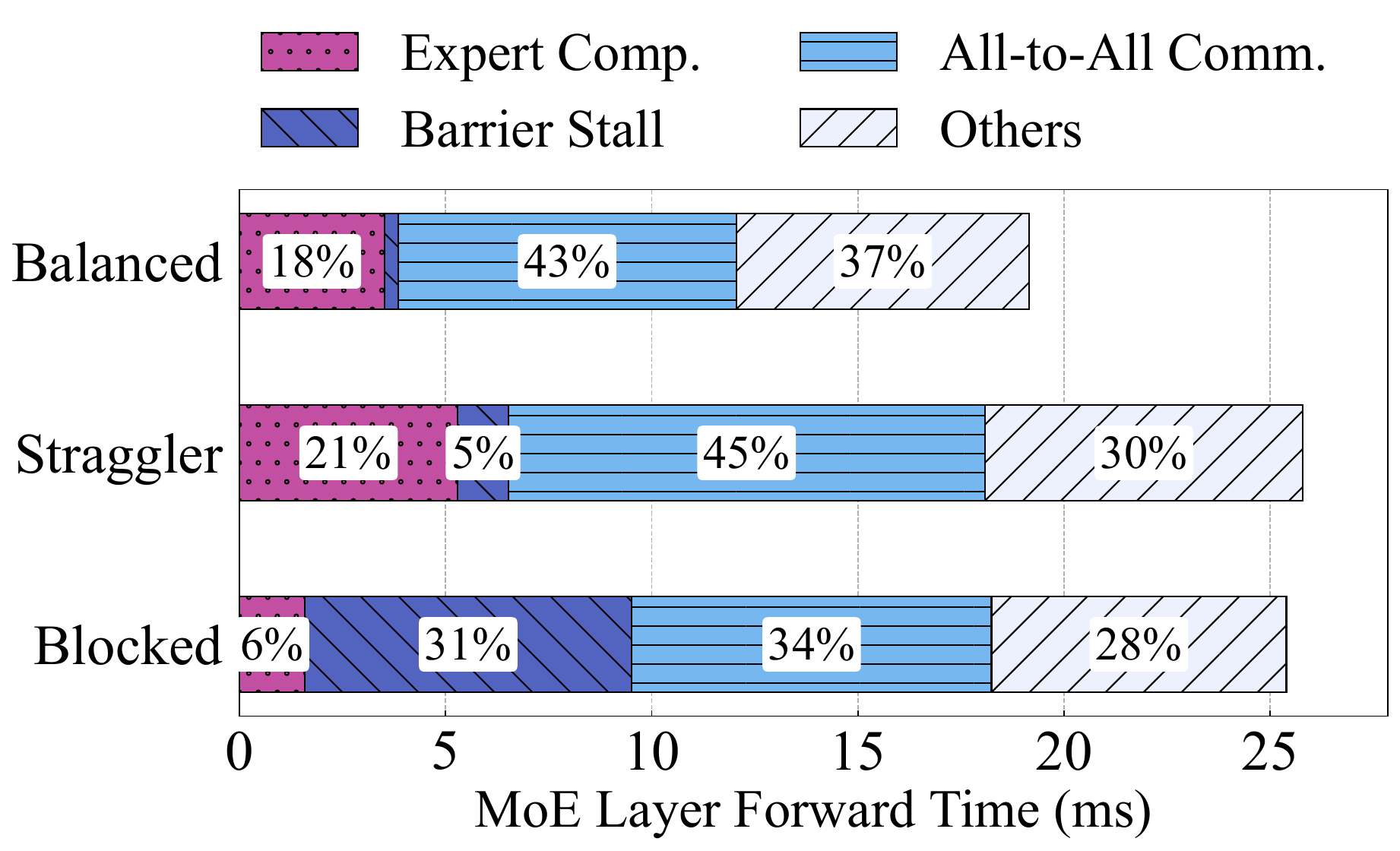}
    \vspace*{-0.1in}
    \caption{One MoE layer's forward time breakdown across different micro-batches.}
    \vspace*{-0.1in}
    \label{fig:background:forward_time_breakdown}
\end{figure}

Figure~\ref{fig:background:forward_time_breakdown} illustrates how fluctuating load imbalance degrades performance. 
The ``Blocked'' and ``Straggler'' cases show the forward time breakdowns of the same MoE 
layer of Qwen3-235B-A22B on the mixed domain, measured on the same GPU for two different micro-batches. 
For comparison, we also evaluate a ``Balanced'' setting that enforces balanced routing. Relative to the 
Balanced setting, the Straggler case incurs substantially longer expert 
computation and communication time, indicating that excessive routed tokens 
directly create stragglers on the corresponding GPUs. 
By contrast, although the Blocked case has shorter local computation time, waiting still accounts for 31\% of its  
total layer forward time because it is stalled by remote stragglers. 
Overall, such fluctuating imbalance patterns increase the forward time to 
1.4$\times$ compared to balanced routing. 
In real deployments, MoE training is typically parallelized with DP, PP, and EP, so such unpredictable 
skew can propagate beyond a single layer, 
creating pipeline bubbles under PP and cross-rack stragglers under DP, further amplifying the 
performance degradation.

\subsection{Opportunity and Challenge}
\label{sec:background:opportunity_and_challenge}

\paraf{Opportunity: routing replay.} As described in Section~\ref{sec:introduction}, 
RL offers a unique load-balancing opportunity that is absent in MoE pretraining. 
Because rollout and training use the same set of MoE parameters to process the same token sequence, 
the token-to-expert routing information of each training sample can be shared between the two stages. 
This enables the training system to observe the exact expert demand of upcoming micro-batches, 
rather than predicting it from recent history, allowing proactive and fine-grained load balancing. 
Recent industrial practice~\cite{ma2025stabilizing, liu2025deepseek-v3-2, zheng2025stabilizing} has also adopted 
routing replay in MoE RL training to improve training stability.
Moreover, RL training includes post-rollout processing before training~\cite{zhong2025optimizing, zhu2026towards}, 
whose time budget is orders of magnitude larger than the runtime decision window in pretraining, 
typically only tens of microseconds for each micro-batch at one layer. 
This additional time budget gives the system substantially more time to exploit routing replay.

\parabf{Challenge: practical load-balancing planning.} Knowing the exact routing of a micro-batch 
does not translate into a practical load-balancing plan. For each MoE layer of each micro-batch, the system must 
decide how to redistribute load through expert reordering, expert replication, or both, while respecting GPU memory 
limits and communication costs. The resulting problem is a large combinatorial optimization that 
recurs throughout training. Even simplified variants are already NP-hard. If expert replication is disabled and memory 
constraints are ignored, the problem reduces to makespan minimization on identical machines 
(\(P||C_{\max}\))~\cite{graham1979optimization}. With replication and capacity constraints, it further contains 
replica placement decisions similar to the facility location problem~\cite{cornuejols1977uncapacitated}. 
Therefore, the key challenge is not merely to exploit routing replay, but to translate it into a practical 
load-balancing plan that remains effective at micro-batch granularity under realistic system constraints and time budgets.

\parabf{Key insight: timescale decomposition.} The key insight is that, under cross-node EP, 
expert reordering and expert replication should operate at different timescales. 
Although both can mitigate load imbalance, expert reordering is poorly suited to micro-batch-level fluctuating imbalance.
Because hot experts may shift across nodes from one micro-batch to the next, 
reordering typically requires coordinated placement changes across the EP group to preserve per-GPU memory balance. 
This makes reordering a global operation whose cross-node coordination cost can outweigh its benefit 
when invoked too frequently. In contrast, expert replication fits the micro-batch timescale better: it adds 
capacity only to hot experts without disturbing the placement of others. It can therefore react quickly to 
transient load spikes using high-bandwidth intra-node links. Accordingly, expert reordering 
should be amortized across training batches for coarse-grained cross-node balancing, while expert replication 
should be applied at micro-batch granularity for fine-grained local adaptation.

%% file: sections/overview.tex
\section{\sysname Overview}
\label{sec:overview}

\begin{figure}[t!]
    \centering
    \includegraphics[width=\linewidth]{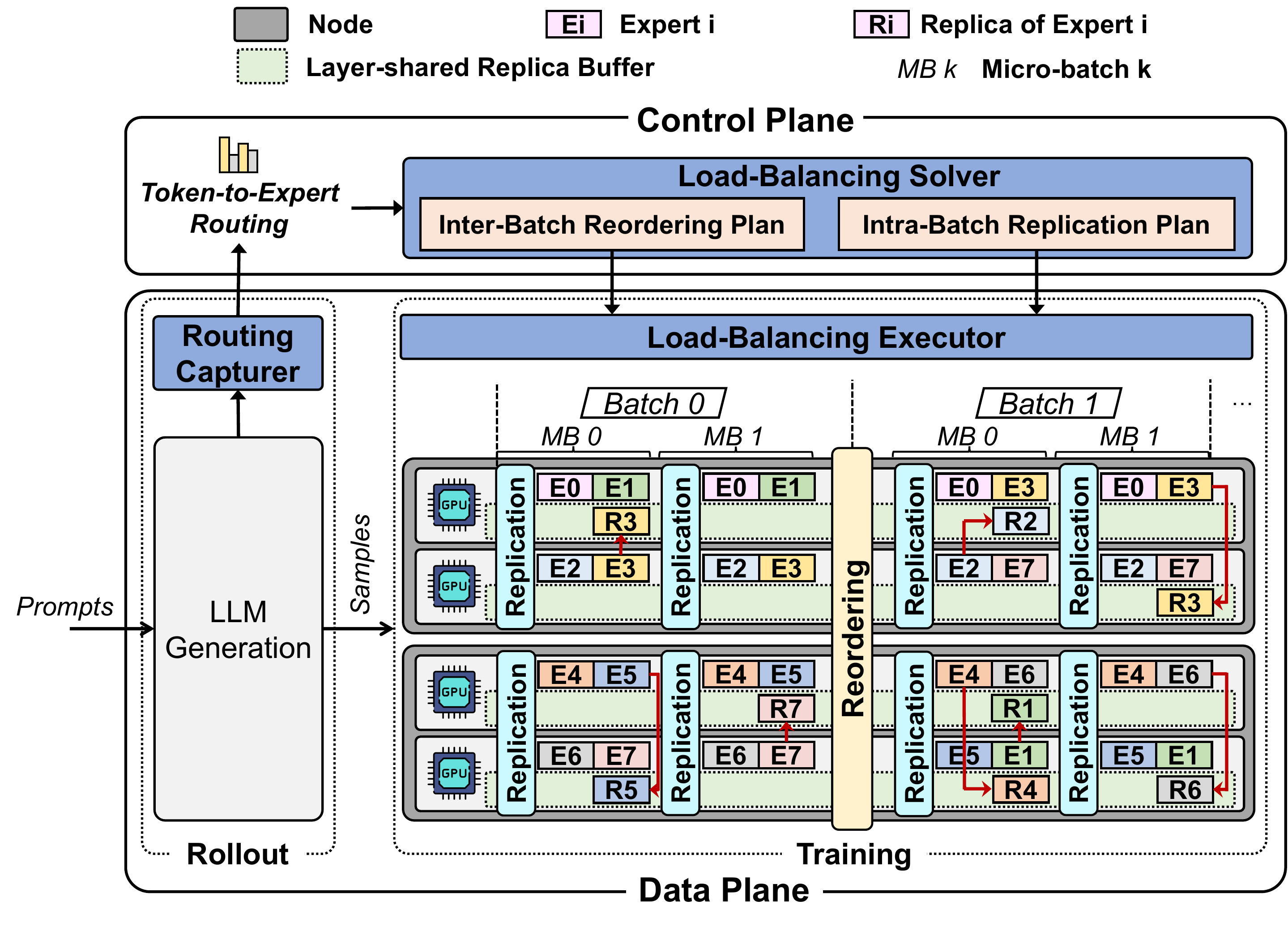}
    \vspace*{-0.2in}
    \caption{Overview of \sysname.}
    \vspace*{-0.1in}
    \label{fig:overview:architecture}
\end{figure}

We propose \sysname, an MoE RL training system that performs fine-grained load balancing at micro-batch granularity 
to address fluctuating load imbalance. Figure~\ref{fig:overview:architecture} shows the overall architecture of \sysname. 

\parabf{Control Plane.} The control plane maintains a global view of the 
RL training workflow. It receives the token-to-expert routing information of 
each sample during rollout and uses it to solve 
fine-grained load-balancing problems. 
Notably, \sysname overlaps this optimization with post-rollout processing, 
such as reward computation and KL-divergence evaluation, 
so its solving time stays off the critical path.

\parait{\uline{Load-Balancing Solver.}} As \sysname places expert reordering and replication 
at inter- and intra-batch timescales, the load-balancing solver decomposes the 
original problem into two sub-problems: inter-batch expert reordering and 
intra-batch expert replication. Inter-batch expert reordering aims to balance GPU loads across the EP group 
over an entire training batch. Intra-batch expert replication determines the replica count,  
placement, and token distribution among replicas for each hot expert. 
Once the solver finishes, the control plane forwards the resulting plans to the data plane for execution.

\parabf{Data Plane.} The data plane is responsible for collecting routing information and 
executing load-balancing decisions.

\parait{\uline{Routing Capturer.}} During rollout, the model generates response tokens in an auto-regressive, 
token-by-token manner. This component records the routed experts for each token at every generation step. 
Once generation is complete, it aggregates the token-level routing decisions for each sample and forwards them 
to the control plane.

\parait{\uline{Load-Balancing Executor.}} The load-balancing executor receives the control plane's plans 
before training begins. For each training batch, it first reorders experts across the EP group for  
coarse-grained cross-node balancing. It then dynamically replicates experts within each node at micro-batch 
granularity to absorb load spikes through high-bandwidth intra-node links. 
To support efficient replication, the executor manages a lightweight, layer-shared replica buffer and 
performs two operations: transferring hot-expert parameters to destination GPUs during the forward pass 
and aggregating replica gradients back during the backward pass. By overlapping communication with 
attention computation and reusing the buffer, the load-balancing executor performs both operations with negligible overhead.

%% file: sections/design.tex
\section{Inter-Batch Expert Reordering}
\label{sec:design:inter-batch}

\sysname optimizes the inter-batch expert reordering plan across the EP group,
typically spanning multiple nodes. In this section, we first formally define the 
inter-batch load-balancing problem. Then we propose a swap-based simulated 
annealing algorithm, which efficiently converges to a balanced solution and 
can further be used to improve data locality.

\subsection{Problem Formulation}
\label{sec:design_inter:formulation}

Inter-batch expert reordering aims to balance the accumulated load over the 
entire training batch according to the aggregated routing information. 
For each MoE layer, \sysname jointly considers loads of both expert computation and 
all-to-all communication to minimize the overall execution time of the 
MoE module. Table~\ref{tab:design:notations} lists the key notations.

\begin{table}[t!]
    \centering
    \resizebox{\linewidth}{!} {
    \begin{tabular}{ll}
        \toprule
        \textbf{Symbol} & \textbf{Description} \\
        \midrule
        $\mathcal{G}$ & Set of GPUs in the EP group \\
        $E, E_g$ & Set of all experts and experts on GPU $g$ \\
        $x_{j,e}$ & Number of tokens routed to expert $e$ from GPU $j$ \\
        $\tau(j,e)$ & Source-destination type of tokens from GPU $j$ to expert $e$ \\
        $y_{j,e,g}$ & Fraction of tokens routed to expert $e$ from GPU $j$ to $g$ \\
        $z_{e,g}$ & Binary indicator of whether $e$'s replica can be placed on $g$ \\
        $L_g^{\mathrm{comp}}$ & GPU $g$'s computation load \\
        $C_g^{\mathrm{nvlink,tx/rx}}$ & GPU $g$'s NVLink send/receive load \\
        $C_g^{\mathrm{rdma,tx/rx}}$ & GPU $g$'s RDMA send/receive load \\
        $T_g^{\mathrm{comp / comm}}$ & MoE computation/communication time on GPU $g$ \\
        $T_{\mathrm{MoE}}$ & Execution time of the MoE module \\
        \bottomrule
    \end{tabular}
    }
    \vspace{-0.05in}
    \caption{Key notations in the design.}
    \vspace{-0.15in}
    \label{tab:design:notations}
\end{table}

\parabf{From token-to-expert routing decisions to computation and communication loads.} 
In line with prior work~\cite{eplb, lplb, zhang2025popfetcher}, we measure load in number of tokens. 
Since each GPU is an execution unit of the MoE module, we first demonstrate 
how to derive the computation and communication loads from the token-to-expert routing.
For each source GPU $j$ and expert $e$, let $x_{j,e}$ denote the number of tokens originally 
located on GPU $j$ (i.e., for attention computation) and routed to expert $e$. 
We assume each expert resides entirely on a single GPU, rather than being sharded across multiple GPUs, 
which is common practice in recent fine-grained MoE models~\cite{liu2024deepseek, team2025kimi, chen2025minimax}.
Let $\mathcal{G}$ be the set of GPUs in the EP group. Let $E_g$ be the set of experts hosted by GPU $g$. 
The total load of GPU $g$ is
\begin{align}
    \label{eq:design:comp_load}
    L_g^{\mathrm{comp}} = \sum_{e \in E_g} l_e = \sum_{e \in E_g} \sum_{j \in \mathcal{G}} x_{j,e}, 
\end{align}
where $l_e$ is the load of expert $e$ hosted by GPU $g$.

\begin{figure}[t!]
    \centering
    \includegraphics[width=0.75\linewidth]{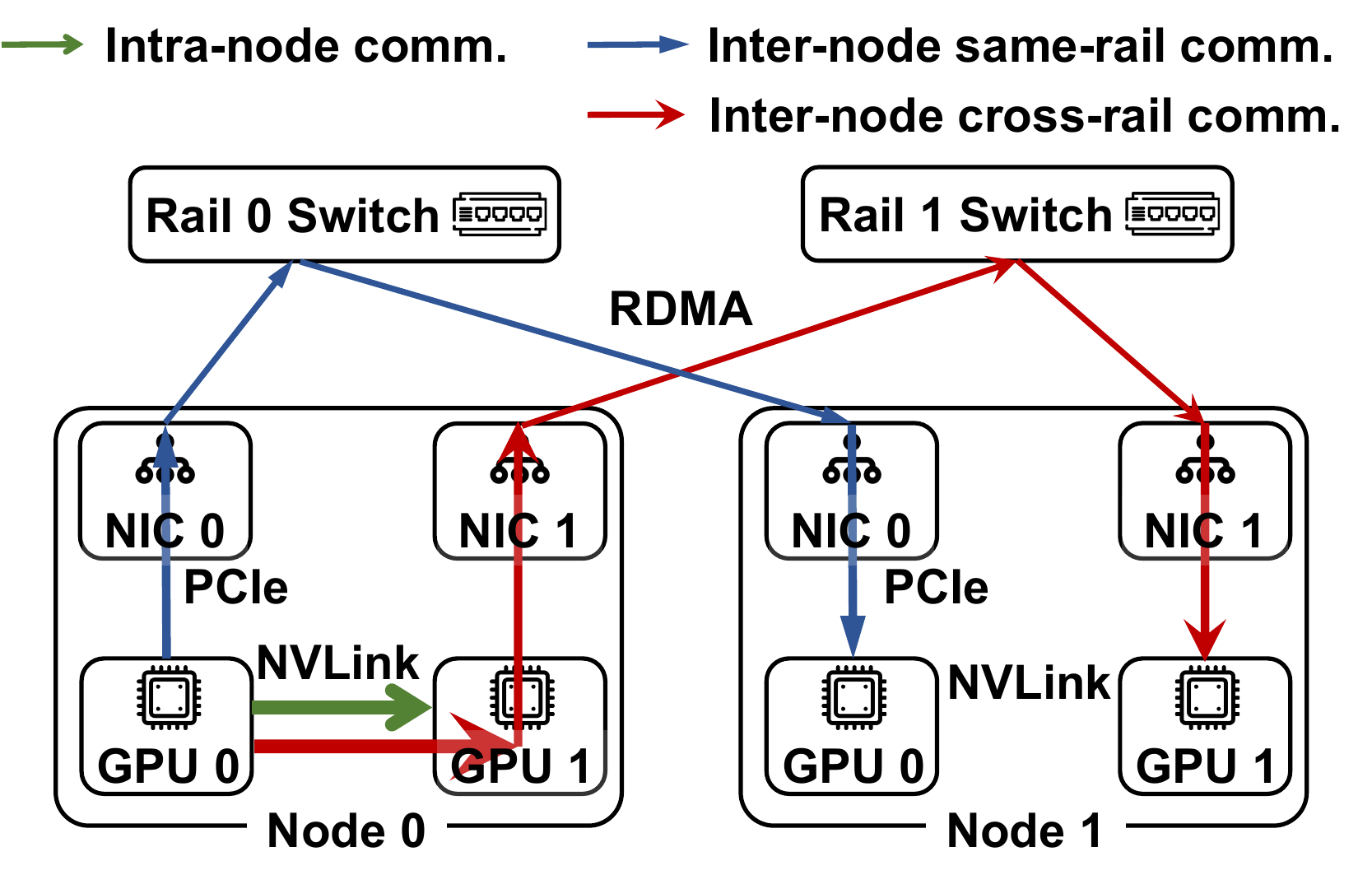}
    \vspace*{-0.1in}
    \caption{Rail-optimized topology for MoE training. The spine switch is omitted for brevity.}
    \vspace*{-0.1in}
    \label{fig:design:rail_optimized}
\end{figure}

Communication load is more complex to characterize. To better support the collective 
communication patterns (e.g., all-reduce) prevalent in LLM training, 
industrial systems have widely adopted rail-optimized network topologies~\cite{qian2024alibaba, nvidia-rail-gtc24}. 
Figure~\ref{fig:design:rail_optimized} shows an example of two nodes with two GPUs each. 
In such a topology, GPUs and NICs with the same index are attached to the same PCIe switch and 
form a rail, and NICs on the same rail across nodes are connected by a leaf switch. 
Consequently, communication in MoE training exhibits three patterns: $(i)$ intra-node communication 
over NVLink, $(ii)$ inter-node same-rail communication with GPUDirect RDMA~\cite{gpu-direct-rdma}, 
and $(iii)$ inter-node cross-rail communication, which first traverses NVLink to reach a GPU on the 
target rail and then uses RDMA for inter-node transfer~\cite{nccl, deepep}.

Given the rail-optimized topology, we classify each routed token by the 
relative location between its source GPU and the GPU hosting its destination expert. 
Let $\tau(j,e)$ denote the source-destination type for tokens on GPU $j$ 
routed to expert $e$, where $\tau(j,e) \in \{\mathrm{loc}, \mathrm{nv}, \mathrm{sr}, \mathrm{cr}\}$ 
corresponds to local memory copy, intra-node NVLink transfer, inter-node 
same-rail RDMA transfer, and inter-node cross-rail transfer with an extra NVLink 
relay, respectively. The source-destination type determines the communication 
load injected into NVLink and RDMA. For GPU $g$, the NVLink load during 
dispatch can be formulated as:
\begin{align}
    \label{eq:design:nvlink_load}
    C_g^{\mathrm{nvlink,tx}} = \sum_{e:\tau(g,e)\in\{\mathrm{nv},\mathrm{cr}\}} x_{g,e}, \\
    C_g^{\mathrm{nvlink,rx}} = \sum_{e \in E_g} \sum_{\substack{j: \\ \tau(j,e)\in\{\mathrm{nv},\mathrm{cr}\}}} x_{j,e},
\end{align}
where $\mathrm{tx}$ and $\mathrm{rx}$ denote the send and receive directions, 
respectively. The $\mathrm{nv}$ and $\mathrm{cr}$ types contribute as direct 
and relayed intra-node traffic, respectively. Similarly, the RDMA load can be 
formulated by replacing the $\mathrm{nv}$ with the $\mathrm{sr}$ type in the 
above equations. For the combine phase, we only need to update each token's 
source GPU to the destination GPU in the dispatch phase, because combine sends 
tokens back to their original locations.

\parabf{From loads to MoE module execution time.} We then use the computation and 
communication loads to estimate the execution time of the MoE module, 
including expert computation and all-to-all communication, for the entire EP group, which we aim to minimize. 
For computation, given the model's hidden size $h$ and intermediate size $h'$, 
processing one token by one expert requires approximately $6hh'$ floating-point 
operations ($2hh'$ for each of the three GEMMs), ignoring element-wise activation and gating overhead for simplicity. Let $F$ denote the effective compute 
capability (FLOPS) of one GPU. The computation time on GPU $g$ is 
\begin{align}
    \label{eq:design:comp_time}
    T_g^{\mathrm{comp}} = \frac{6hh'L_g^{\mathrm{comp}}}{F}.
\end{align}
For communication, given the effective bandwidth of NVLink and RDMA as 
$B_{\mathrm{nvlink}}$ and $B_{\mathrm{rdma}}$, respectively, the communication 
time on GPU $g$ is 
\begin{align}
    \label{eq:design:comm_time}
    T_g^{\mathrm{comm}} = \max_{\phi \in \{\mathrm{tx}, \mathrm{rx}\}} \left\{ \max \left\{ \frac{C_g^{\mathrm{nvlink, \phi}}}{B_{\mathrm{nvlink}}}, \frac{C_g^{\mathrm{rdma, \phi}}}{B_{\mathrm{rdma}}} \right\} \right\}.
\end{align}
The outer maximum is due to the full-duplex nature of communication links, and 
the inner maximum is because existing all-to-all implementations~\cite{nccl, deepep} 
optimize cross-node communication efficiency by pipelining and overlapping NVLink 
and RDMA transmissions. 
\sysname obtains $F$, $B_{\mathrm{nvlink}}$, and $B_{\mathrm{rdma}}$ values through profiling and 
interpolation as prior work does~\cite{zhong2024distserve, zhu2025megascale, zhang2025disttrain, zhang2026heddle}.
Therefore, the MoE execution time of the EP group is 
\begin{align}
    \label{eq:design:moe_time}
    T_{\mathrm{MoE}} = \max_{g \in \mathcal{G}} T_g^{\mathrm{comp}} + \max_{g \in \mathcal{G}} T_g^{\mathrm{comm}}.
\end{align}

\begin{algorithm}[t!]
    \small
    \caption{Swap-based simulated annealing}
    \label{alg:design:sa}
    \begin{algorithmic}[1]
        \Require Experts $E$, GPUs $\mathcal{G}$, routing matrix $\{x_{j,e}\}$, seed set $\mathcal{S}$, cooling rate $\gamma$, termination threshold $\epsilon$
        \Ensure Expert reordering plan $\Pi^\star$
        \State $\Pi_0 \gets \Call{GreedyLPT}{\{x_{j,e}\}, \mathcal{G}}$  \Comment{initial state}
        \ForAll{$s \in \mathcal{S}$ \textbf{in parallel}}
            \State $\Pi \gets \Pi_0$; $\Pi_s^\star \gets \Pi_0$
            \State $T \gets \widetilde{T}_{\mathrm{MoE}}(\Pi)$; $T_s^\star \gets T; \theta \gets T$ 
            \While{$\theta > \epsilon$}
                \State $(e_a, e_b) \gets \Call{RandomSample}{E}$
                \State $\Pi' \gets \Call{Swap}{\Pi, e_a, e_b}$  \Comment{incremental state update}
                \State $T' \gets \widetilde{T}_{\mathrm{MoE}}(\Pi')$  \Comment{objective smoothing}
                \State $\Delta \gets T - T'$
                \If{$T' < T$ \textbf{or} $\Call{Rand}{0,1} < e^{-\Delta / \theta}$}
                    \State $\Pi \gets \Pi'$; $T \gets T'$
                \EndIf
                \If{$T < T_s^\star$}
                    \State $\Pi_s^\star \gets \Pi$; $T_s^\star \gets T$
                \EndIf
                \State $\theta \gets \gamma \theta$
            \EndWhile
        \EndFor
        \State \Return $\arg\min_{\Pi \in \{\Pi_s^\star \mid s \in \mathcal{S}\}} T_{\mathrm{MoE}}(\Pi)$
    \end{algorithmic}
\end{algorithm}
\vspace{-0.2in}

\subsection{Swap-based Simulated Annealing}
\label{sec:design:inter-batch:sa}

An exhaustive search over expert reordering plans has a search space of $O(|E|!)$. 
Recent MoE models~\cite{yang2025qwen3, liu2024deepseek} typically contain $>100$ (e.g., 128 or 256) experts per layer, 
making the problem combinatorially intractable. 
A natural solution is a greedy longest-processing-time (LPT)~\cite{graham1969bounds} heuristic, which  
sorts experts in descending order of computation load ($l_e$) and places each expert on the  
GPU satisfying the capacity constraint and having the smallest current total load. 
However, the greedy method lacks a global perspective. 
Reordering an expert changes not only its computation load contribution, 
but also the relative locations between that expert and its routed tokens, 
thereby altering the communication volume and imbalance across GPUs. 
Thus, the greedy approach does not yield optimal performance, as we show in \S~\ref{sec:evaluation:ablation}. 

Instead, we propose a swap-based simulated annealing algorithm that 
jointly considers computation and communication loads by building on Formula~\ref{eq:design:moe_time}, 
searching for a better solution. 
For each MoE layer, the algorithm applies the LPT heuristic as an initial state. 
It then repeatedly swaps two randomly chosen experts, and reevaluates the MoE execution 
time $T_{\mathrm{MoE}}$. A new state is always accepted if it improves the objective and is 
otherwise accepted with a temperature-controlled probability, allowing the search to 
escape local minima. We set the initial temperature to the initial state's objective value. 
We launch multiple annealing runs with different random seeds in parallel and keep the 
best final reordering plan. Algorithm~\ref{alg:design:sa} shows the pseudocode.

\parabf{Incremental state update.} The swap-based search naturally enables an incremental
update scheme, which significantly reduces the cost of evaluating each swap. When two
experts are exchanged, only the routing contributions associated with these two experts
can change. Therefore, instead of recomputing all GPU loads from scratch, \sysname
rescans $\{x_{j,e}\}$ only for the swapped experts and incrementally updates the affected
computation and communication loads. The time complexity of this incremental update is $O(|\mathcal{G}|)$.

\parabf{Objective smoothing.} The exact objective $T_{\mathrm{MoE}}$ defined above contains nested $\max$ operators, 
which create large plateaus in the search landscape: a swap may improve several near-bottleneck GPUs but leave the 
current maximum unchanged. To make the objective continuous and differentiable, we replace each $\max$ with 
a log-sum-exp (LSE) surrogate: $\operatorname{LSE}_\beta (z_1, \ldots, z_n) = \frac{1}{\beta} \log \sum_{k=1}^{n} e^{\beta z_k}$, 
where a larger $\beta$ yields a closer approximation to $\max$. 
We then smooth the objective function as 
$\widetilde{T}_{\mathrm{MoE}} = \operatorname{LSE}_\beta \left(\widetilde{T_g}^{\mathrm{comp}}, g \in \mathcal{G} \right) + \operatorname{LSE}_\beta \left( \widetilde{T_g}^{\mathrm{comm}}, g \in \mathcal{G} \right)$ 
and smooth the inner $\max$ of communication time as well. 
This smoothed objective provides more optimization signals. Swaps that reduce
near-critical loads can still decrease $\widetilde{T}_{\mathrm{MoE}}$ even when the exact
maximum does not change. We compare the performance of different $\beta$ values and 
choose $\beta = 20$ for \sysname. 

\parabf{Optimizing data locality.}
Cross-node EP is often combined with data parallel attention~\cite{liu2024deepseek} to
increase the batch size and reduce TP communication. Under this setup, each GPU
hosts a replica of the attention module, so each sample can in principle be processed by
any GPU. This flexibility creates an additional optimization opportunity: assigning a
sample to a GPU that hosts more of its frequently activated experts improves locality and
reduces MoE communication loads.

Given an expert reordering plan $\Pi^\star$, \sysname performs a second round of
swap-based simulated annealing to optimize sample placement. This stage uses the same
objective as expert reordering, but replaces expert swaps with sample swaps. Swapping two
samples changes their source GPUs and therefore changes the resulting communication load
and communication time. The final output is a locality-aware sample-to-GPU assignment that
further reduces cross-GPU MoE traffic.

\section{Intra-Batch Expert Replication}
\label{sec:design:intra-batch}

At the intra-batch timescale, \sysname dynamically replicates experts within a node 
based on the inter-batch expert reordering plan to absorb micro-batch-level load 
fluctuations. This design poses two challenges. On the systems side, expert replication 
must operate at micro-batch granularity without incurring excessive memory or 
synchronization overhead. On the algorithmic side, once an expert has one or more 
replicas, the system must determine the number of replicas, their placement, 
and the token assignment among them. Existing solutions~\cite{eplb, zhang2025popfetcher} 
use fixed replication plans at global-batch granularity and thus sidestep 
these challenges. In this section, we address these challenges with 
a holistic design for efficient intra-batch expert replication.

\subsection{Layer-Shared Replica Buffer}
\label{sec:design:intra-batch:replica_buffer}

\begin{figure}[t!]
    \centering
    \includegraphics[width=\linewidth]{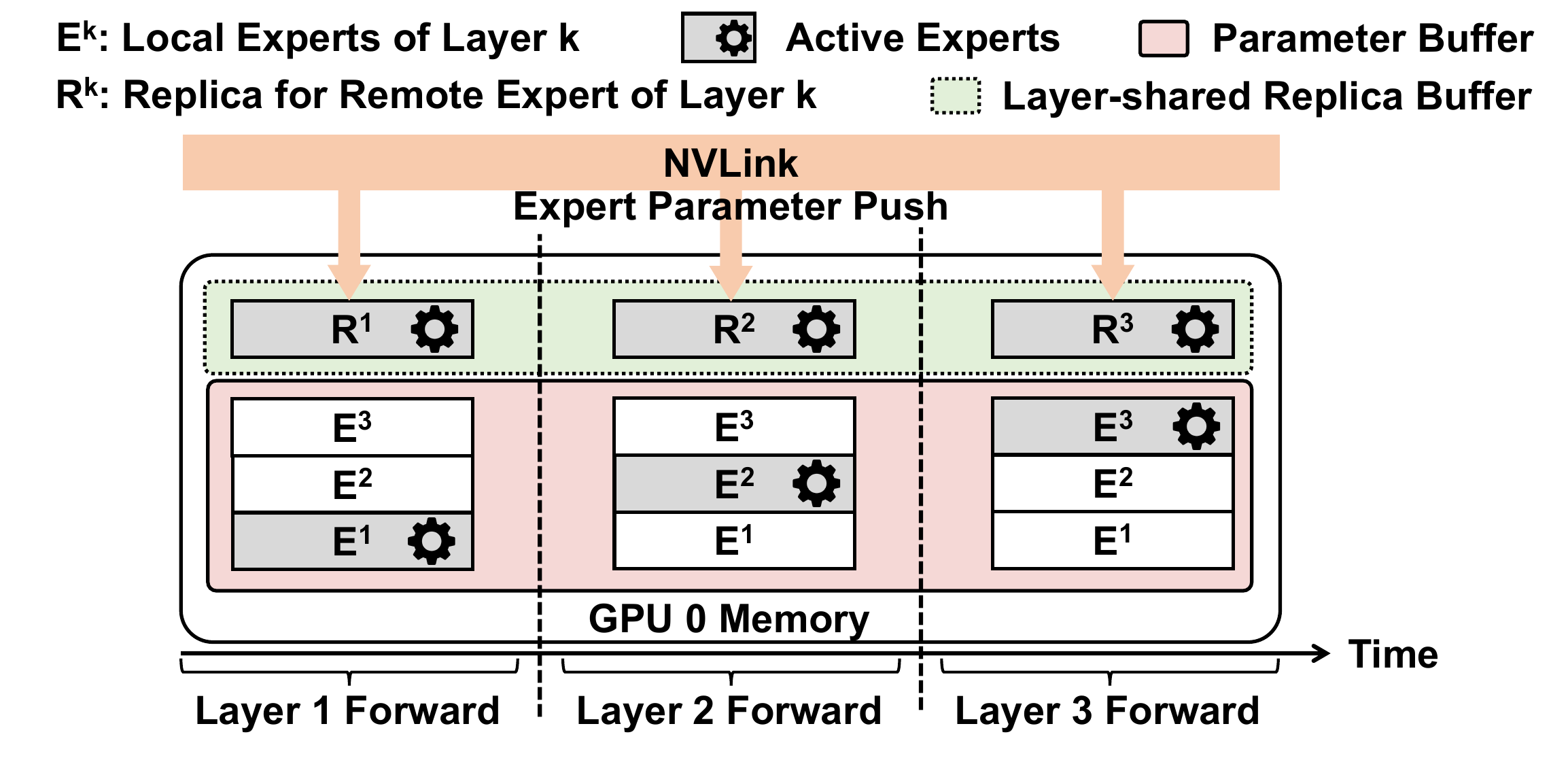}
    \vspace*{-0.2in}
    \caption{Layer-shared replica buffer for fine-grained intra-batch replication.}
    \vspace*{-0.1in}
    \label{fig:design:replica_buffer}
\end{figure}

A straightforward replication design would reserve replica slots for every MoE layer, so that
replicas can be materialized immediately whenever a layer is executed. 
However, the forward pass proceeds layer by layer, and only the current layer's expert parameters 
are accessed at any given time. 
\sysname exploits this property with a layer-shared replica buffer: each GPU 
allocates only a small number of replica slots, stores replicas for one layer 
at a time, and reuses the same buffer across all layers.
Figure~\ref{fig:design:replica_buffer} illustrates this design with an example 
where each GPU can host at most one expert replica. As the forward pass 
progresses from Layer~1 to Layer~3, the resident experts of all layers remain 
in place, while the replica buffer is reused to hold the replicas required 
by each layer in turn. Specifically, the buffer is filled with the 
replicas needed by Layer~1 before expert computation, then overwritten with those for 
Layer~2, and so on for subsequent layers.

\parabf{Memory overhead.} 
The key advantage of a layer-shared buffer is its negligible memory overhead. 
Suppose each GPU stores $M$ experts under EP in each of $L$ MoE layers, and up to $r$ replicas may be
placed on the GPU for any single layer. A naive design that reserves a separate replica
buffer for every layer requires $Lr$ replica slots in total. In contrast,
\sysname allocates only one shared buffer with $r$ slots and reuses it across all
layers, reducing the additional memory overhead by one to two orders of magnitude for 
common MoE models with tens of layers. Since expert parameters dominate replica memory consumption, 
this reduction makes micro-batch-level replication practical for deep MoE models.

\parabf{Synchronization overhead.}
Leveraging routing replay, \sysname makes load-balancing decisions before the current micro-batch is executed, 
determining which experts need to be replicated for each layer and on which GPUs these replicas should be placed. 
At runtime, \sysname uses a push-based SPMD scheme to populate the shared replica buffer. 
During the forward pass, each GPU independently pushes local experts to their required destination GPUs 
before the replicas are actually used. During the backward pass, the gradients produced by the replicas of this 
micro-batch are pushed back and accumulated on the source GPUs. Because replication is confined within a node, 
this synchronization uses only high-bandwidth intra-node links. \sysname further overlaps the synchronization operations
with attention computation, avoiding any interference with the critical path. 

\subsection{Micro-batch-level Replication Planning}
\label{sec:design:intra-batch:planning}

\paraf{MILP formulation.} For each expert $e$ of an MoE layer in a given micro-batch, 
\sysname constructs a replication plan with two coupled components. The first component is 
\uline{replica placement}: determining whether $e$ should be replicated and, if so, 
on which GPUs within the local node the replicas should be placed. 
The second component is \uline{token splitting}: determining how the tokens routed 
to $e$ should be distributed across the original expert and its replicas. 
These two components must be optimized jointly. Replica placement determines the available locations 
for routed tokens, while token splitting determines whether a placement actually reduces the 
peak computation and communication loads. 

We now formulate micro-batch-level replication planning for one MoE layer of one micro-batch
as a mixed-integer linear program. The input is the routing matrix $\{x_{j,e}\}$ after
inter-batch expert reordering. Let $\mathcal{G}(e) \subseteq \mathcal{G}$ denote the
GPUs in the same node as expert $e$'s home GPU, i.e., the candidate GPUs on which replicas of
$e$ may be placed. We introduce a binary placement variable $z_{e,g} \in \{0,1\}$, where
$z_{e,g}=1$ means tokens of expert $e$ are allowed to be served by GPU $g$. For the home GPU
that already hosts $e$, $z_{e,g}$ is fixed to $1$; for other candidate GPUs, $z_{e,g}$ represents the
replica placement decision. We also introduce a continuous variable
$y_{j,e,g} \in [0, 1]$, denoting the fraction of tokens on source GPU $j$ that are routed
to expert $e$ and finally served by GPU $g$. Under this definition, the actual number of tokens
assigned from source GPU $j$ to GPU $g$ for expert $e$ is $x_{j,e} y_{j,e,g}$. 
Using the same time symbols as in the inter-batch reordering problem for the MoE 
execution time of the micro-batch, the intra-batch replication planning problem is formulated as:
\begin{align}
    \min_{y_{j,e,g} \in [0, 1], \enspace z_{e,g} \in \{0, 1\}} T_{\mathrm{MoE}} & = T^{\mathrm{comp}}(y, z) + T^{\mathrm{comm}}(y, z), \label{eq:design:replication_objective} \\
    \text{s.t.} \quad \sum_{g \in \mathcal{G}(e)} y_{j,e,g} &= 1, \quad \forall j, e \text{ with } x_{j,e}>0, \label{eq:design:replication_conservation} \\
    y_{j,e,g} &\le z_{e,g}, \quad \forall j, e, g \in \mathcal{G}(e), \label{eq:design:replication_feasibility} \\
    z_{e,g} &= 1, \quad \forall g, \forall e \in E_g, \label{eq:design:replication_home} \\
    \sum_{e \in E \setminus E_g: g \in \mathcal{G}(e)} z_{e,g} &\le r, \quad \forall g, \label{eq:design:replication_capacity}
\end{align}
where Constraint~\ref{eq:design:replication_conservation} enforces that the token fractions for
each non-empty routing entry $(j,e)$ sum to $1$, so the split induced by $y_{j,e,g}$ fully
partitions the original token count $x_{j,e}$. 
This relation also allows any objective originally expressed using routed-token counts to be
rewritten with the split variables $x_{j,e} y_{j,e,g}$, and hence as a function of the decision
variables $y_{j,e,g}$ and $z_{e,g}$.
Constraint~\ref{eq:design:replication_feasibility} couples token splitting with replica placement, 
ensuring that a positive fraction can be assigned to GPU $g$ only when the original expert or one 
of its replicas is present there. Constraint~\ref{eq:design:replication_home} fixes the placement of 
the home GPU. Constraint~\ref{eq:design:replication_capacity} captures the per-GPU replica buffer limit 
introduced in \S~\ref{sec:design:intra-batch:replica_buffer}. Since the objective contains only linear terms 
and $\max$ operators with respect to the decision variables, it can be converted into an MILP problem 
by introducing auxiliary variables to linearize the maxima.

\parabf{Solving MILP.}
Although the MILP formulation is compact, solving it for every micro-batch remains 
computationally expensive. This difficulty arises from two sources. First, replica 
placement introduces binary variables over expert-GPU pairs, causing the search 
space to grow combinatorially with the number of candidate experts and available 
replica slots. Second, general-purpose MILP solvers such as SCIP~\cite{bolusani2024scip} incur 
substantial branch-and-bound overhead and often fail to produce high-quality 
solutions within the time budget (e.g., millisecond-level per micro-batch) in our setting.

\begin{algorithm}[t!]
    \small
    \caption{Incremental greedy heuristic}
    \label{alg:design:greedy_ilp}
    \begin{algorithmic}[1]
        \Require Experts $E$ and reordering plan $\Pi$, GPUs $\mathcal{G}$, routing matrix $\{x_{j,e}\}$, candidate GPU sets $\{\mathcal{G}(e)\}$, replica slot limit $r$
        \Ensure Replica placement plan $z^\star$ and token splitting plan $y^\star$
        \State $z_{e,g} \gets 1$ iff $e \in E_g$, otherwise $0$
        \State $y_{j,e,g} \gets 0$ for all $j, e, g$  \Comment{initialize without replica}
        \While{$\exists g$ with available replica slot}
            \State $g_b \gets \arg\max_g T_{MoE, g}(\Pi, y, z)$  \Comment{bottleneck GPU}
            \State $e^\star \gets \arg\max_e T_{g, e}(E_g, y, z)$  \Comment{bottleneck expert}
            \State $g_t \gets \arg\min_{g \in \mathcal{G}(e^\star) \land z_{e^\star,g}=0 \land \Call{HasSlot}{g}} T_{MoE, g}(\Pi, y, z)$  
            \State \Comment{least-load GPU}
            \If{no feasible $g_t$ exists}
                \State \textbf{break}
            \EndIf
            \State $z_{e^\star,g_t} \gets 1$
            \State $y \gets \Call{SolveTokenSplitLP}{\{x_{j,e}\}, z}$
        \EndWhile
        \State \Return $(z, y)$
    \end{algorithmic}
\end{algorithm}

We propose an incremental greedy heuristic to avoid direct MILP solving. 
The heuristic incrementally constructs the replica placement and re-optimizes 
token splitting after each placement decision. 
Algorithm~\ref{alg:design:greedy_ilp} shows the detailed procedure.
It starts from the no-replica state and then optimizes the replication plan in an 
iterative manner. In each iteration, it first identifies the current bottleneck GPU $g_b$ 
using computation and communication loads. Among the experts on $g_b$, it selects 
the one contributing the largest load. Next, one replica is added to the feasible GPU with 
the smallest current load. Finally, given a new replica placement $z$, it re-solves the 
token-splitting subproblem using linear programming to obtain the optimal split ratios $y$. 
This process repeats until no replica slot remains.

%% file: sections/implementation.tex
\section{Implementation}
\label{sec:implementation}

We implement \sysname based on Slime~\cite{slime} for RL workflow orchestration and Megatron-LM~\cite{megatron-lm} for MoE training 
with 14K lines of Python, C++, and CUDA. We use SGLang~\cite{zheng2024sglang} as the rollout backend. 

\parabf{Routing capturer.} Built on top of SGLang, the routing capturer exposes routing information 
after each response generation. We store the routing metadata in a small GPU buffer (tens of MB) and 
asynchronously transfer it to host memory, incurring negligible overhead during rollout.

\parabf{Load-balancing solver.} We implement both the inter-batch reordering and intra-batch replication solvers in C++ with 
OpenMP~\cite{openmp}. Each thread solves one independent optimization task: one layer with one simulated-annealing seed for 
inter-batch reordering, or one layer with one micro-batch for intra-batch replication. 

\parabf{Layer-shared replica buffer.} We implement layer-shared replica buffer synchronization with a persistent 
asynchronous TMA kernel~\cite{tma} based on the NVIDIA Hopper architecture. To minimize interference with attention computation, 
we limit the synchronization kernel to occupy only a small number of streaming processors (SMs), leaving most SMs available 
for compute-intensive attention and thereby overlapping synchronization with computation while minimizing contention.

%% file: sections/evaluation.tex
\section{Evaluation}
\label{sec:evaluation}

In this section, we first compare the overall performance of \sysname against 
state-of-the-art MoE training systems (\S~\ref{sec:evaluation:overall}). 
To further validate \sysname's effectiveness, we strengthen history-based 
load-balancing methods by providing oracle loads from routing replay. 
We then perform ablation studies to quantify the contribution of each component 
of \sysname and compare them with strawman baselines (\S~\ref{sec:evaluation:ablation}). 
Next, we demonstrate the near-optimal imbalance reduction capability of \sysname (\S~\ref{sec:evaluation:case_study}). 
Finally, we investigate the system overhead of \sysname (\S~\ref{sec:evaluation:overhead}).

\parabf{Testbed.} We conduct experiments on a 20-node cluster, where each node has eight NVIDIA Hopper GPUs 
and 1.5 TB of host memory. Nodes are connected by a rail-optimized InfiniBand RDMA network 
with an aggregate bandwidth of 8$\times$400 Gbps per node, while intra-node GPU communication is facilitated 
by 900 GB/s NVLink. Our software stack includes PyTorch 2.8.0, CUDA 12.6 (driver 535.161.08), 
FlashAttention-3~\cite{shah2024flashattention}, TransformerEngine 2.7.0~\cite{transformer-engine}, 
and DeepEP 1.1.0~\cite{deepep}. The load-balancing solver in \S~\ref{sec:overview} runs on a separate CPU-only node 
with 128 Intel Xeon CPU cores, and we use Ray~\cite{moritz2018ray} to dispatch solving tasks.

\begin{table}[t!]
    \centering
    \arrayrulewidth=0.5pt
    \extrarowheight=1pt
    \resizebox{0.9\linewidth}{!} {
        \begin{tabular}{@{}cccccc@{}}
            \arrayrulecolor{black}\hline
            \arrayrulecolor{black}\hline
            Model             & \#MoE Layers & \#GPUs & DP & PP & EP \\ \hline
            Qwen3-30B-A3B     & 48 & 64 & 4 & 1 & 16 \\
            GLM4.5-106B-A12B  & 45 & 64 & 1 & 2 & 32 \\
            Qwen3-235B-A22B   & 94 & 160 & 1 & 5 & 32 \\ 
            \arrayrulecolor{black}\hline
            \arrayrulecolor{black}\hline
        \end{tabular}
    }
    \vspace{-0.1in}
    \caption{Parallelism strategies for different models.}
    \vspace{-0.15in}
    \label{tab:evaluation:parallelism}
\end{table}

\parabf{Workloads and models.} We evaluate \sysname on three representative RL workloads: reasoning, 
instruction following, and coding. The corresponding datasets and rollout hyperparameters are the same as those 
used in Table~\ref{tab:background:datasets} in \S~\ref{sec:background}. We also mix these datasets to better 
reflect production RL training workloads, as suggested by DeepSeek-V3.2~\cite{liu2025deepseek-v3-2}. 
We evaluate \sysname on Qwen3-30B-A3B~\cite{yang2025qwen3}, GLM4.5-106B-A12B~\cite{zeng2025glm}, 
and Qwen3-235B-A22B~\cite{yang2025qwen3}, which span diverse model scales and come from different vendors. 
These models have 128 experts per layer and choose top-$8$ experts for each token.

\parabf{Baselines.} We compare \sysname with the following baselines. 
\begin{itemize}[leftmargin=*]
    \item \textbf{Megatron-LM}~\cite{megatron-lm} is a widely used, open-source LLM training 
    framework that supports hybrid parallelism and integrates various MoE optimizations, including DeepEP~\cite{deepep} 
    and GroupedGEMM for efficient all-to-all communication and expert computation, respectively.
    \item \textbf{PopFetcher}~\cite{zhang2025popfetcher} predicts which experts in the current layer are most likely to 
    be activated for the current training batch based on recent cross-layer activation patterns, then prefetches them for 
    replication and overlaps the prefetching with attention computation. We tune its window size to achieve the 
    best performance in our experiments.
    \item \textbf{EPLB$^{+}$} strengthens EPLB~\cite{eplb} by replacing historical load observations with 
    oracle loads from routing replay. EPLB, proposed by DeepSeek-V3~\cite{liu2024deepseek}, is a history-based MoE 
    load-balancing method that uses greedy algorithms to optimize expert reordering and replica placement at global-batch granularity, 
    considering only computation load. 
    \item \textbf{LPLB$^{+}$} strengthens LPLB~\cite{lplb} in the same way as EPLB$^{+}$ by providing oracle loads. 
    LPLB is an EPLB variant with dynamic token splitting for each micro-batch via a GPU-based linear programming solver. 
    To reduce solving overhead, it restricts each expert to at most one replica. 
    \item \textbf{Balanced} is a system-level ideal baseline that randomly routes tokens to 
    all experts to achieve optimal load balancing. However, it does not preserve the model's original routing decisions and 
    thus serves only as a reference for the best-case system efficiency.
\end{itemize}
For a fair comparison, we integrate all methods into Megatron-LM and use the same parallelism strategy. 
We further employ the same layer-shared replication buffer, the number of replication slots 
(two replica slots per GPU in our experiments), and the synchronization overlapping 
mechanism for all baselines using expert replication. This ensures that the performance gains of \sysname arise from its 
load-balancing techniques rather than other system optimizations.

\parabf{Metrics.} \sysname targets the fluctuating load-imbalance problem during training and is orthogonal to 
rollout optimizations. We therefore use training throughput, defined as the throughput of the training stage, 
as our primary metric. In the case study (\S~\ref{sec:evaluation:case_study}), we also report rank-level skewness, 
defined as the ratio of the maximum to the average number of tokens processed across GPUs, to quantify the 
degree of load imbalance.

\begin{figure}[t!]
    \centering
    \includegraphics[width=\linewidth]{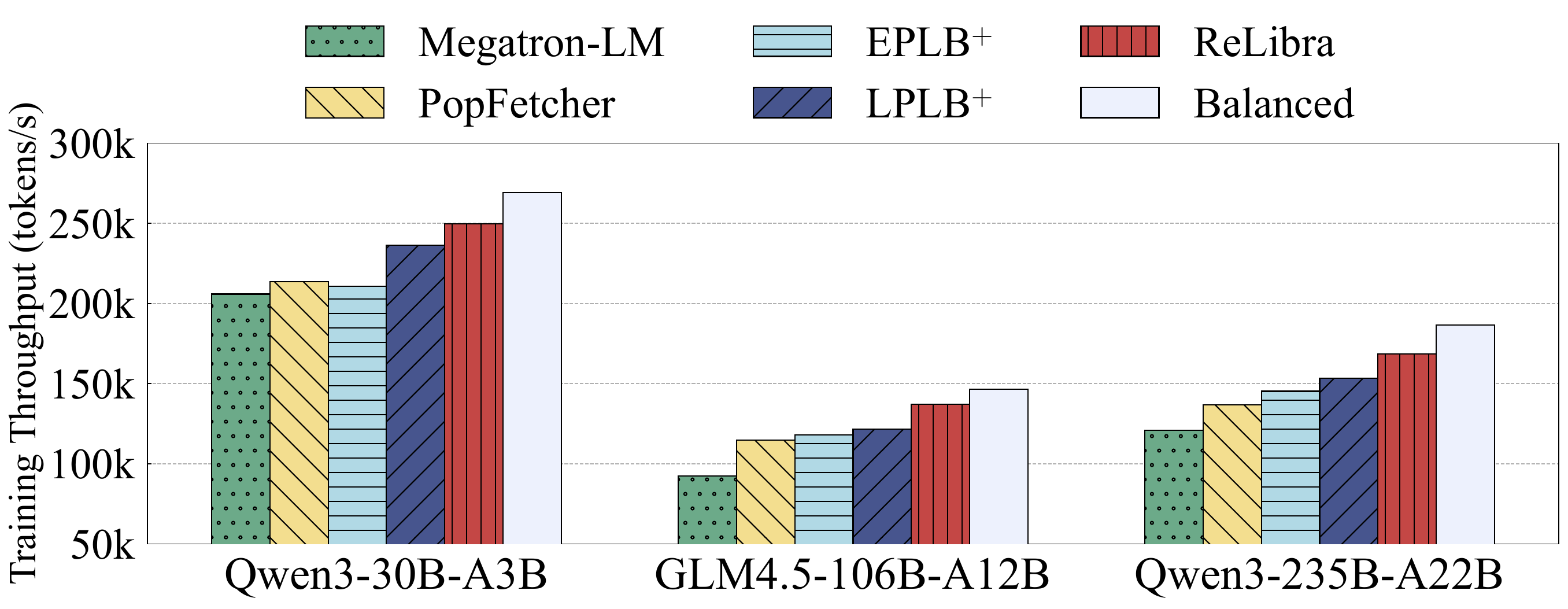}
    \vspace*{-0.2in}
    \caption{Training throughput on the mixed dataset with different models.}
    \vspace*{-0.1in}
    \label{fig:evaluation:overall_different_models}
\end{figure}

\begin{figure}[t!]
    \centering
    \includegraphics[width=\linewidth]{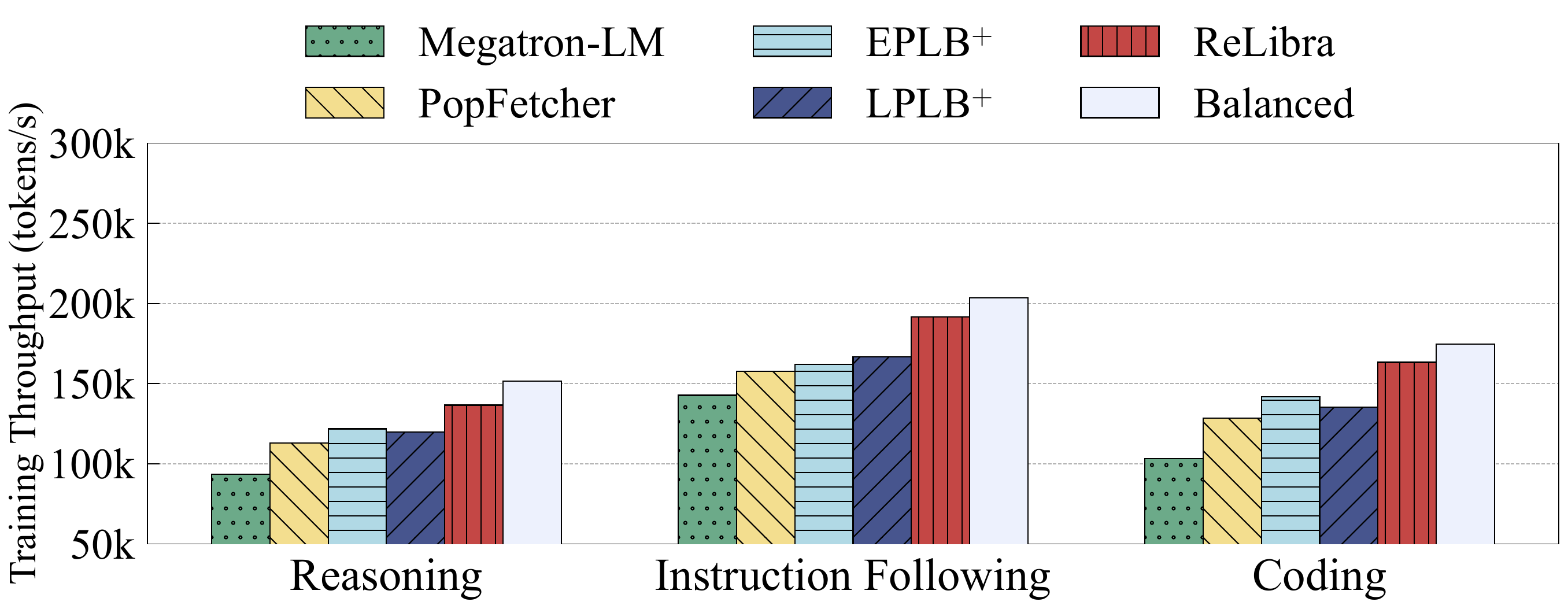}
    \vspace*{-0.2in}
    \caption{Training throughput on different datasets with Qwen3-235B-A22B.}
    \vspace*{-0.1in}
    \label{fig:evaluation:overall_different_datasets}
\end{figure}

\subsection{Overall Performance}
\label{sec:evaluation:overall}

We first benchmark \sysname against baselines on different models and datasets. 
We deploy the three models with different parallelism strategies based on their parameter scales. 
Specifically, we set the EP size to 16 for the 30B model and 32 for the other two models, and we 
employ data parallel attention as industrial practice suggests~\cite{liu2024deepseek} to reduce TP communication overhead. 
Table~\ref{tab:evaluation:parallelism} details the configurations. 
We apply the same global batch size, 1024, for all methods under all settings. 
Figure~\ref{fig:evaluation:overall_different_models} and Figure~\ref{fig:evaluation:overall_different_datasets} 
present the average training throughput across various models and datasets. We summarize the results as follows.
\begin{itemize}[leftmargin=*]
    \item \sysname outperforms all baselines, achieving speedups of 1.21--1.58$\times$ over Megatron-LM, 1.17--1.27$\times$ over PopFetcher, 
    and 1.06--1.21$\times$ over EPLB$^{+}$ and LPLB$^{+}$, respectively. The gains come from \sysname's ability to capture 
    fluctuating load imbalance at micro-batch granularity and adapt its load-balancing decisions accordingly. In contrast, 
    PopFetcher depends on imperfect load prediction, while EPLB$^{+}$ and LPLB$^{+}$ use static expert reordering and replication plans 
    throughout the training batch. As a result, even with oracle loads, they cannot respond to fine-grained imbalance fluctuations.
    \item \sysname achieves comparable performance to Balanced, reaching 90\%-94\% of its training throughput. This is enabled by its 
    hierarchical design, which uses inter-batch expert reordering for coarse-grained balancing and intra-batch expert 
    replication for fine-grained adaptation, leading to near-balanced execution for each micro-batch.
    \item \sysname is effective across RL domains, model scales, and parallelism strategies. LPLB$^{+}$ performs slightly worse 
    than EPLB$^{+}$ on reasoning and coding, but better on instruction following and mixed workloads. This is because LPLB$^{+}$ 
    uses linear programming to dynamically split tokens across replicas, while restricting each expert to at most one replica to 
    keep the search space manageable. As a result, it is more effective for highly fluctuating imbalance but less effective for 
    strongly skewed imbalance. \sysname avoids this trade-off by replaying routing and computing replication plans before training, 
    while overlapping solver execution with post-rollout processing so that it adds no extra training time.
\end{itemize}

\begin{table}[t!]
    \centering
    \arrayrulewidth=0.5pt
    \extrarowheight=1pt
    \resizebox{\linewidth}{!} {
        \begin{tabular}{c|c|c|c|c|c}
            \arrayrulecolor{black}\hline
            \arrayrulecolor{black}\hline
            \multirow{3}{*}{\begin{tabular}[c]{@{}c@{}} Idx \end{tabular}} & \multirow{3}{*}{\begin{tabular}[c]{@{}c@{}} Method \end{tabular}} & \multicolumn{2}{c|}{Reasoning} & \multicolumn{2}{c}{Mixed} \\ 
            \cline{3-6} 
             & & Normalized & \multirow{2}{*}{$\Delta$} & Normalized & \multirow{2}{*}{$\Delta$} \\ 
             & & Throughput &  & Throughput &  \\ 
            \hline
            1 & Megatron-LM & 1  &  & 1 & \\ 
            2 & (1) with inter-batch reordering & 1.12  & +12\% & 1.15  & +15\% \\ 
            3 & (2) with intra-batch replication & 1.46  & +34\%  & 1.39  & +24\% \\ 
            \arrayrulecolor{black}\hline
            \arrayrulecolor{black}\hline
        \end{tabular}
    }
    \vspace{-0.1in}
    \caption{Throughput improvement breakdown of inter-batch and intra-batch load balancing when training Qwen3-235B-A22B.}
    \vspace{-0.1in}
    \label{tab:evaluation:ablation_contributions}
\end{table}

\subsection{Ablation Study}
\label{sec:evaluation:ablation}

In this subsection, we perform ablation studies to quantify the contribution and verify the effectiveness of each 
component of \sysname. We use Qwen3-235B-A22B as the test model and evaluate it on the reasoning and mixed datasets. 

We first perform an ablation study that incrementally enables inter-batch and intra-batch load-balancing 
techniques to isolate their contributions. Table~\ref{tab:evaluation:ablation_contributions} reports the 
resulting training throughput breakdown. Enabling only inter-batch expert reordering yields speedups of 12\% and 15\% 
on the reasoning and mixed datasets, respectively. Adding intra-batch expert replication further improves throughput 
by 34\% and 24\%, respectively. This is because inter-batch reordering distributes hot experts more evenly across nodes, 
thereby creating more opportunities for effective intra-node replication to absorb micro-batch-level load fluctuations. 
We then deep dive into each component of \sysname. 

\begin{figure}[t!]
    \centering
    \includegraphics[width=0.7\linewidth]{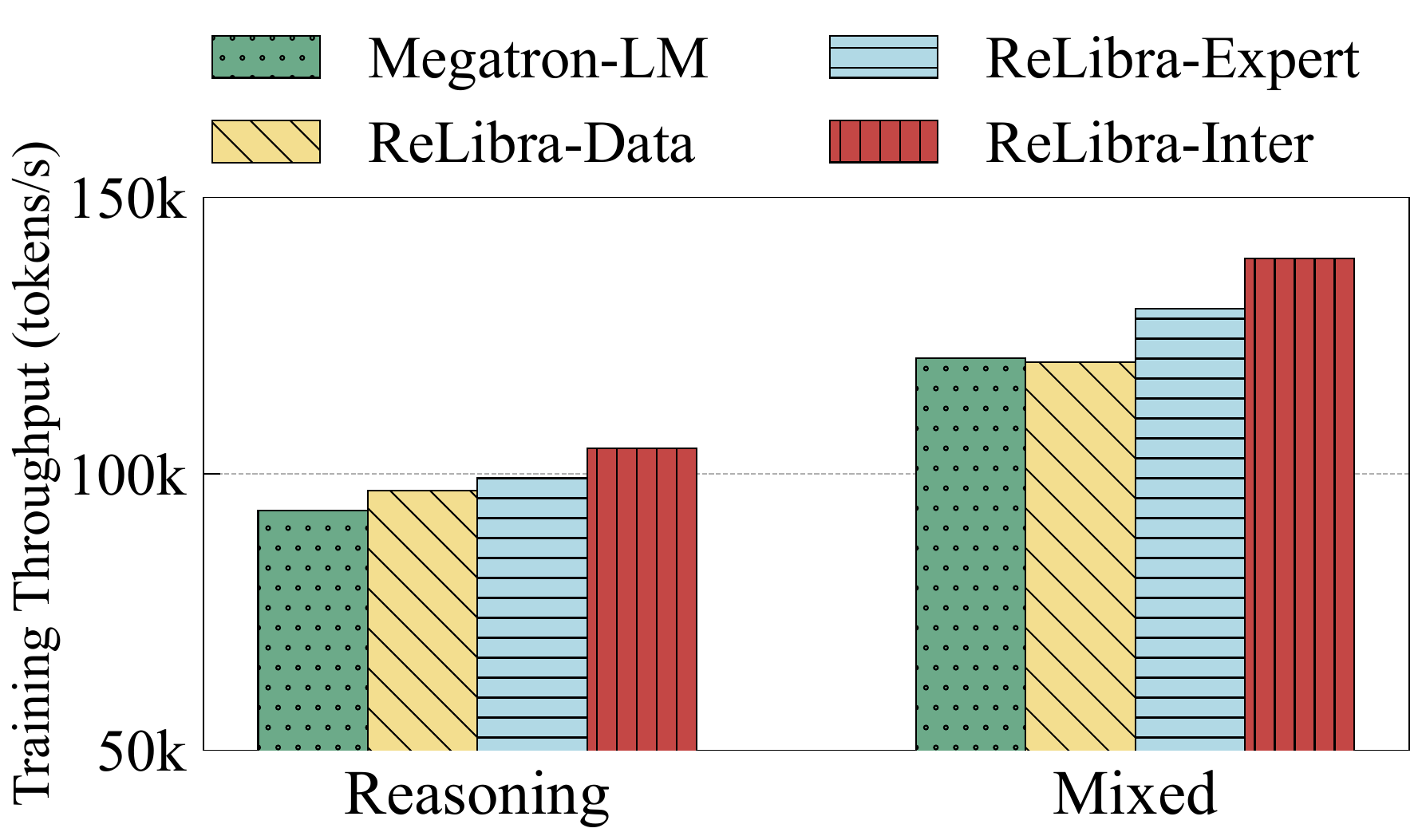}
    \vspace*{-0.1in}
    \caption{Performance of inter-batch expert reordering.}
    \vspace*{-0.1in}
    \label{fig:evaluation:ablation_reordering}
\end{figure}

\parabf{Inter-batch expert reordering.} To evaluate \sysname's inter-batch expert reordering, we 
compare it (referred to as \sysname-Inter) against two greedy reordering baselines: 
\begin{itemize}[leftmargin=*]
    \item \textbf{\sysname-Expert} uses the LPT algorithm to place experts on GPUs, providing the initial state for our 
    swap-based simulated annealing algorithm.
    \item \textbf{\sysname-Data} uses the LPT algorithm to assign data samples to GPUs for data locality optimization. 
    Specifically, it greedily assigns samples from long to short to the GPU with the lowest estimated MoE communication 
    load according to Formula~\ref{eq:design:comm_time}. This assignment serves as the initial solution for 
    the second round of swap-based simulated annealing for data-locality optimization.
\end{itemize}
Figure~\ref{fig:evaluation:ablation_reordering} shows that \sysname-Inter achieves 1.05--1.16$\times$ higher training 
throughput than both baselines, demonstrating its effectiveness in balancing load across GPUs while improving data locality. 
The results also show that although greedy LPT improves performance over the standard EP plan in Megatron-LM, 
our swap-based simulated annealing algorithm converges to better solutions. 

\begin{figure}[t!]
    \centering
    \includegraphics[width=0.7\linewidth]{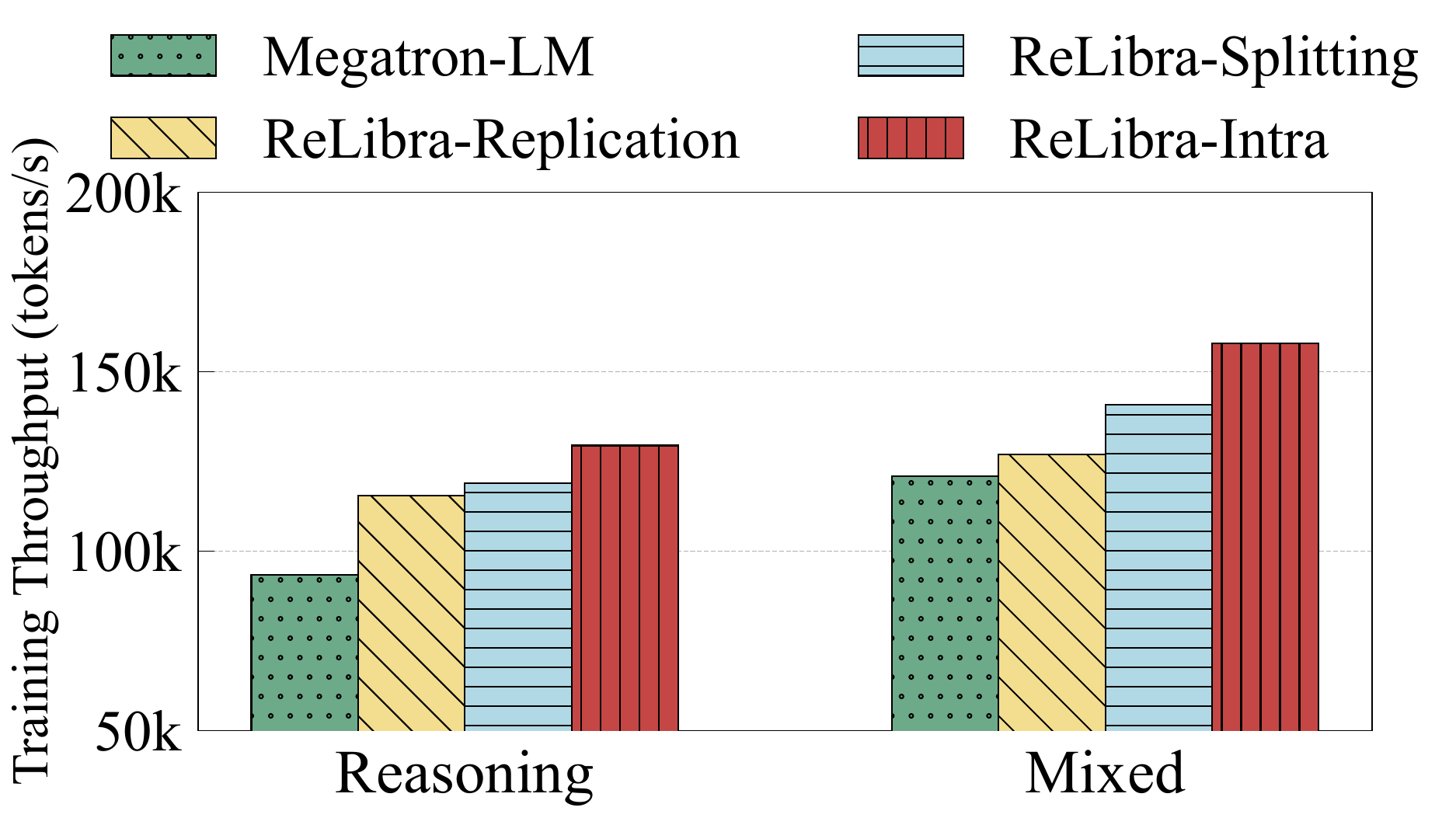}
    \vspace*{-0.1in}
    \caption{Performance of intra-batch expert replication.}
    \vspace*{-0.1in}
    \label{fig:evaluation:ablation_replication}
\end{figure}

\parabf{Intra-batch expert replication.} To evaluate \sysname's intra-batch expert replication, we 
compare it (referred to as \sysname-Intra) against two oracle baselines, each optimizing one component 
of the replication plan, namely replica placement or token splitting:
\begin{itemize}[leftmargin=*]
    \item \textbf{\sysname-Replication} re-optimizes replica placement for each micro-batch using 
    EPLB$^{+}$'s greedy replication algorithm with oracle loads, while retaining EPLB$^{+}$'s round-robin 
    token splitting.
    \item \textbf{\sysname-Splitting} re-optimizes token splitting for each micro-batch by removing 
    the replica count limit in $LPLB^{+}$ and solving offline linear programs with oracle loads.
\end{itemize}
Figure~\ref{fig:evaluation:ablation_replication} shows that \sysname-Intra improves training throughput 
by 1.12--1.24$\times$ over \sysname-Replication and by 1.09--1.12$\times$ over \sysname-Splitting, 
demonstrating the benefit of formulating replication planning as a holistic MILP problem. 
\sysname-Splitting consistently outperforms \sysname-Replication because token splitting provides 
finer-grained control over load distribution. Unlike replica placement, which adjusts capacity 
only in discrete steps, token splitting directly redistributes tokens across available replicas and 
can therefore adapt more precisely to fluctuating load imbalance.

\subsection{Case Study}
\label{sec:evaluation:case_study}

\begin{figure}[t!]
    \centering
    \includegraphics[width=\linewidth]{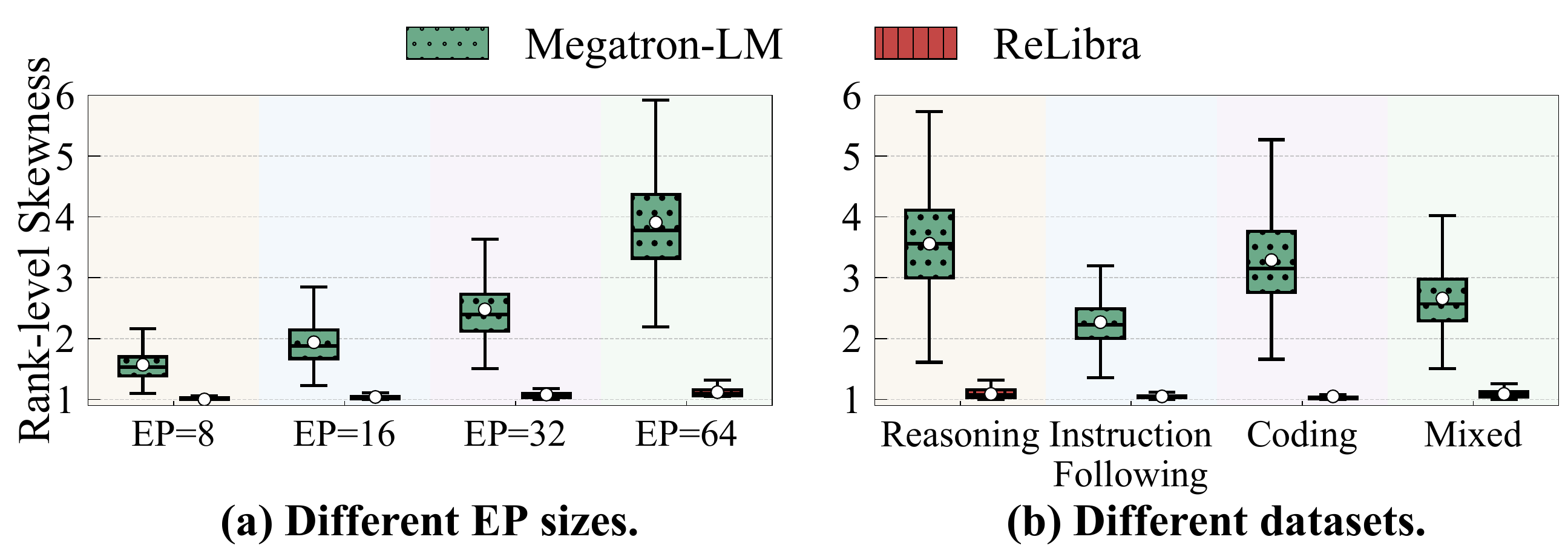}
    \vspace*{-0.2in}
    \caption{Rank-level skewness under different EP sizes and datasets.}
    \vspace*{-0.1in}
    \label{fig:evaluation:skewness}
\end{figure} 

In this subsection, we evaluate the imbalance reduction capability of \sysname across different EP 
sizes and datasets. Due to resource constraints, we use Qwen3-30B-A3B for the EP-size study. 
Figure~\ref{fig:evaluation:skewness}\textcolor{green!80!black}{(a)} shows the distribution of rank-level 
skewness for EP sizes from 8 to 64, where skewness is defined as the ratio of the maximum to the average 
MoE token count across GPUs. According to Formula~\ref{eq:design:comp_load} and \ref{eq:design:nvlink_load}, this metric directly reflects the degree of load imbalance. 
As EP size increases, Megatron-LM exhibits more severe imbalance because each GPU hosts fewer experts, 
making the load distribution less even. In contrast, \sysname maintains an average skewness close to 1.0 
(perfectly balanced) across all EP sizes, with values of 1.00, 1.03, 1.06, and 1.08, respectively. 
Figure~\ref{fig:evaluation:skewness}\textcolor{green!80!black}{(b)} further reports skewness on 
Qwen3-235B-A22B across different datasets, where \sysname achieves average values of 1.02--1.07, 
demonstrating its effectiveness across RL domains with diverse imbalance patterns.

\begin{figure}[t!]
    \centering
    \includegraphics[width=0.75\linewidth]{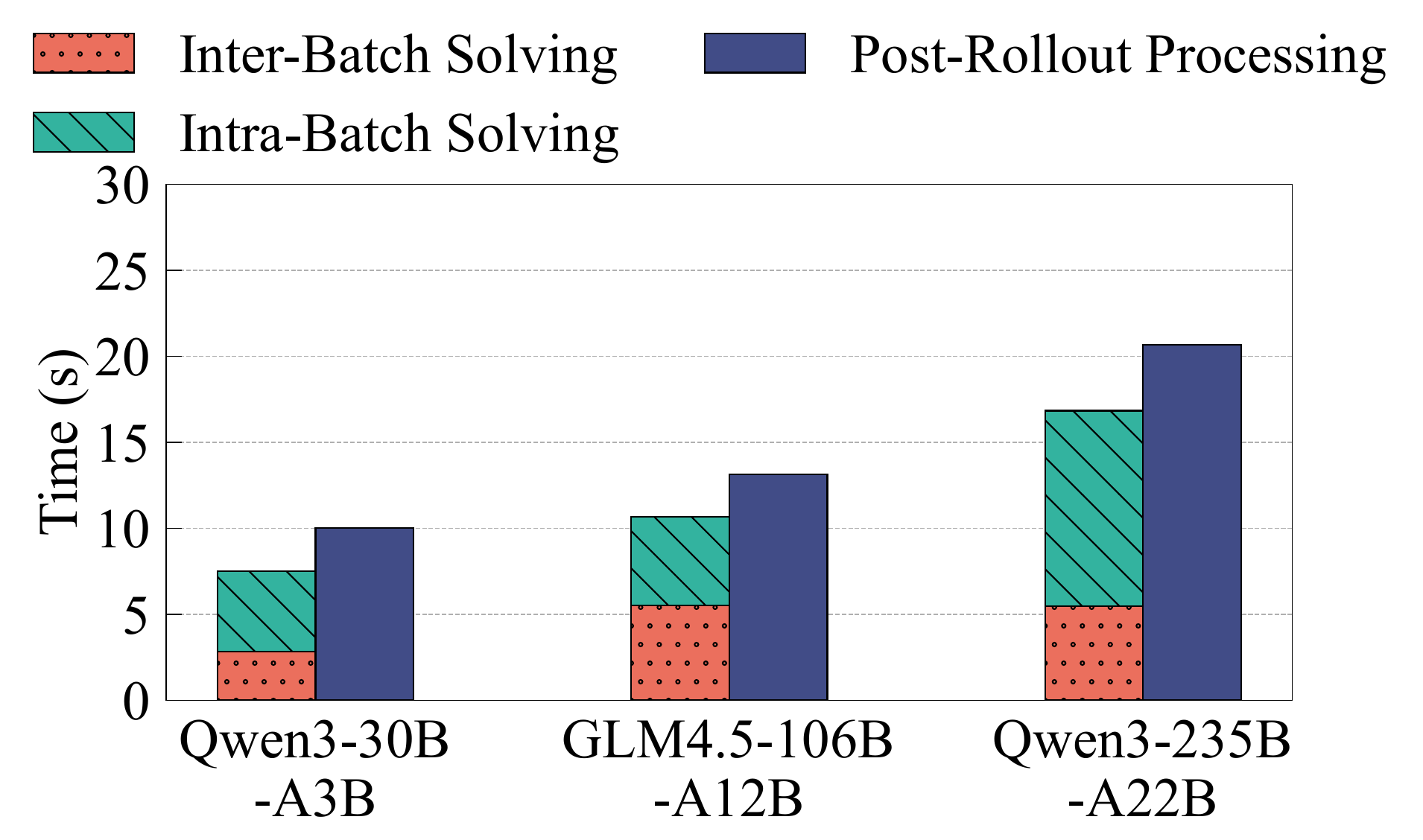}
    \vspace*{-0.1in}
    \caption{Solving time breakdown of \sysname.}
    \vspace*{-0.1in}
    \label{fig:evaluation:overhead_solving}
\end{figure}

\subsection{System Overhead}
\label{sec:evaluation:overhead}

\parabf{Solving time.} We first measure the solving time of \sysname's inter-batch and intra-batch load-balancing solvers, 
which run on a separate CPU node during post-rollout processing before training starts. 
Figure~\ref{fig:evaluation:overhead_solving} shows that the total solving time is fully hidden behind post-rollout processing, 
and thus does not lie on the critical path of RL training for all three models. 
Qwen3-235B-A22B incurs longer intra-batch solving time than the other two models because it contains more MoE layers. 
Since optimization problems are independent across layers, this overhead can be further reduced with more CPU cores 
and higher parallelism. 

\begin{table}[t!]
    \centering
    \arrayrulewidth=0.5pt
    \extrarowheight=1pt
    \resizebox{\linewidth}{!} {
        \begin{tabular}{@{}cccccc@{}}
            \arrayrulecolor{black}\hline
            \arrayrulecolor{black}\hline
            Model             & Layer-shared Replica Buffer & Model Parameter & Ratio \\ \hline
            Qwen3-30B-A3B     & 18 MiB & 6.26 GiB & 0.28\% \\
            GLM4.5-106B-A12B  & 22 MiB & 16.64 GiB & 0.13\% \\
            Qwen3-235B-A22B   & 72 MiB & 4.35 GiB & 1.62\% \\ 
            \arrayrulecolor{black}\hline
            \arrayrulecolor{black}\hline
        \end{tabular}
    }
    \vspace{-0.1in}
    \caption{Additional parameter memory usage of the layer-shared replica buffer per GPU.}
    \vspace{-0.15in}
    \label{tab:evaluation:memory_usage}
\end{table}

\parabf{Expert reordering overhead.} We then measure the overhead of inter-batch expert reordering. 
Because expert reordering operates at the inter-batch timescale, its cost is amortized over the entire training batch. 
With a global batch size of 1024, the average expert reordering time accounts for only 1.4\%--2.1\% of the 
execution time of the training batch across the three models, indicating negligible overhead.

\parabf{Expert replication overhead.} Finally, we measure the overhead of intra-batch expert replication in two aspects:
$(i)$ the additional memory usage of the layer-shared replica buffer, and $(ii)$ the synchronization overhead for dynamic 
expert replication. Table~\ref{tab:evaluation:memory_usage} reports the total parameter memory per GPU and the additional 
memory per GPU used for expert replication in the layer-shared replica buffer for all models. As expected, the extra memory 
is negligible, accounting for only 0.13\%--1.62\% of the total parameter memory. The additional gradient memory is identical, 
since each replicated parameter corresponds to one gradient. This small overhead arises because \sysname stores replicas 
for only one layer at a time and reuses the same buffer across layers. 

\begin{figure}[t!]
    \centering
    \includegraphics[width=\linewidth]{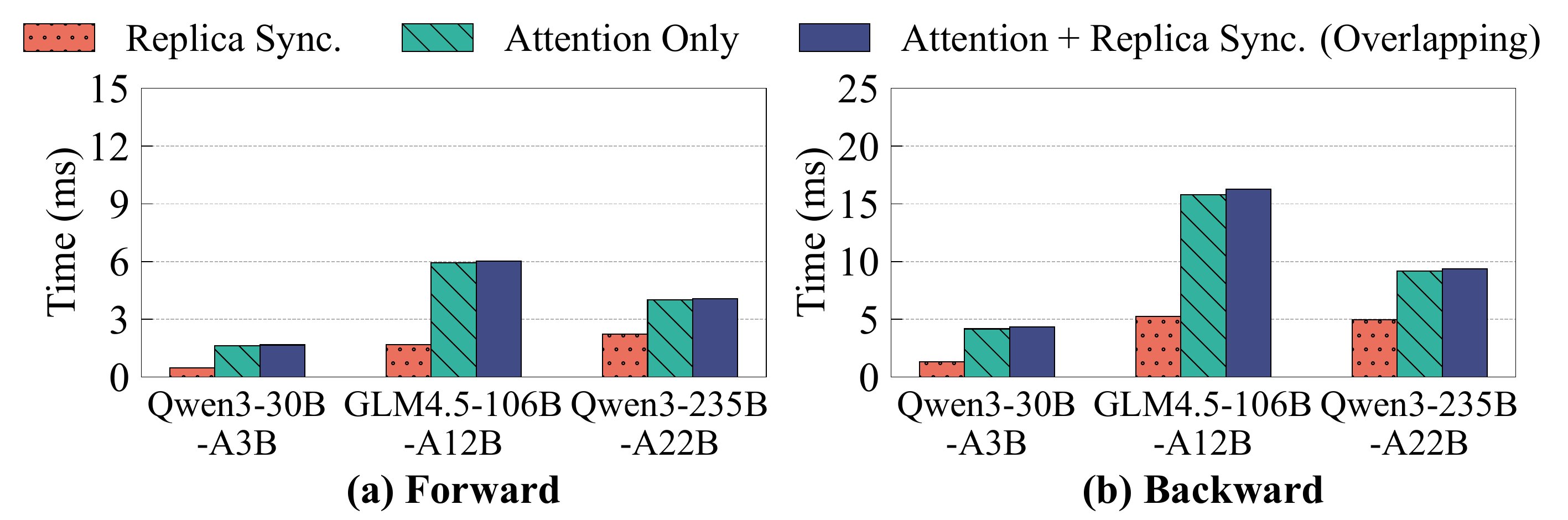}
    \vspace*{-0.2in}
    \caption{Synchronization overhead of \sysname.}
    \vspace*{-0.1in}
    \label{fig:evaluation:overhead_overlapping}
\end{figure}

Figure~\ref{fig:evaluation:overhead_overlapping} shows the synchronization 
overhead of dynamic expert replication during forward and backward. 
In the forward pass, each GPU pushes local experts to the GPUs hosting their 
replicas; in the backward pass, it sends the replica gradients back. 
We refer to this parameter and gradient transfer time as replica synchronization 
time. The figure reports, for all models, the average replica synchronization 
time, the attention computation time without overlap, and the attention 
computation time when overlapped with replica synchronization. 
The results show that replica synchronization is fully hidden behind 
attention computation and introduces negligible interference to its execution 
time. This is because \sysname limits the synchronization kernel to occupy 
only a small number of SMs, leaving most SMs available for attention.

%% file: sections/discussion.tex
\section{Discussion}
\label{sec:discussion}

\paraf{Off-policy RL.} To improve data efficiency and accelerate the RL iteration, some off-policy RL algorithms either reuse 
the same batch of data for multiple parameter updates~\cite{mnih2015human, haarnoja2018soft} or allow rollout and training to 
proceed asynchronously~\cite{team2025kimi-k1.5, zhong2025streamrl}. In these cases, rollout and training do not necessarily 
use the same model parameters, resulting in potentially different routing results. 
However, recent industrial practice~\cite{ma2025stabilizing, liu2025deepseek-v3-2, zheng2025stabilizing} 
has also adopted routing replay in off-policy RL training, which substantially helps stabilize training by guaranteeing that 
experts are trained by the exact tokens generated by themselves.

%% file: sections/related.tex
\section{Related Work}
\label{sec:related}

\parabf{MoE training.} MoE scales model capacity efficiently~\cite{liu2024deepseek, lepikhin2021gshard, rajbhandari2022deepspeed, fedus2022switch} 
but faces challenges in load balancing and all-to-all communication.
Existing MoE load-balancing methods, mostly designed for pretraining, 
use historical observations to predict future expert loads~\cite{eplb, zhang2025popfetcher, li2023accelerating, nie2023flexmoe, zhai2023smartmoe}, 
making them less effective for RL training with fluctuating imbalance. 
Furthermore, the tight decision window in pretraining limits their strategy space~\cite{liu2026laer, zhao2025micromoe}. 
\sysname addresses these problems by using routing replay to obtain exact routing information, 
enabling fine-grained load balancing via expert reordering and replication.
For all-to-all communication, existing solutions mainly follow three directions. 
One line of work~\cite{hwang2023tutel, nie2022hetumoe, deepep} uses hierarchical all-to-all to 
reduce cross-node traffic by exploiting bandwidth asymmetry. Another improves 
efficiency by overlapping communication with computation~\cite{liu2024deepseek, jin2025megascale, chen2024centauri, jiang2024megascale} 
or fusing data movement with compute operators~\cite{zhang2025comet, chang2024flux}. 
A third takes a data-centric approach by moving experts closer to the 
data~\cite{liu2023janus, zhang2025popfetcher}, thereby reducing communication volume when 
experts are small. \sysname is orthogonal to these approaches and can be combined with 
them by adapting Formula~\ref{eq:design:comm_time} to their communication patterns.

\parabf{RL training.} RL training is becoming increasingly important for 
enhancing LLM capabilities and human alignment. 
Existing systems generally adopt two deployment patterns: synchronous RL, where rollout 
and training execute on a shared GPU pool in turn~\cite{sheng2025hybridflow}, and 
asynchronous RL, which overlaps these stages across separate GPU pools to maximize hardware 
utilization~\cite{team2025kimi, zhong2025streamrl, gao2025rollart} but at the cost of training stale samples.
In the synchronous setting, optimizations like Seer~\cite{qin2025seer} and 
TLT~\cite{hu2026taming} accelerate rollout via speculative decoding~\cite{leviathan2023fast, jin2025efficient}, 
while RLHFuse~\cite{zhong2025optimizing} minimizes GPU idle time by initiating training early 
during rollout. 
\sysname is orthogonal to these efforts. While prior works focus on accelerating rollout or 
orchestrating RL workflows, \sysname optimizes the training stage itself. 
\sysname generalizes to both deployment patterns through routing replay, consistent with 
recent industrial approaches for stabilizing MoE training~\cite{ma2025stabilizing, liu2025deepseek-v3-2, zheng2025stabilizing}.

%% file: sections/conclusion.tex
\section{Conclusion}
\label{sec:conclusion}

We present \sysname, an MoE RL training system that addresses fluctuating load imbalance by exploiting routing replay in the RL workflow. 
\sysname decomposes load balancing into inter- and intra-batch timescales, matching expert reordering and replication to 
the hierarchical bandwidths of training clusters. With efficient solving algorithms and a lightweight layer-shared replica 
buffer, \sysname can adapt to load imbalance at micro-batch granularity with negligible overhead. Our evaluation shows 
that \sysname achieves near-optimal load balancing and improves training throughput by up to 1.6$\times$ over 
state-of-the-art MoE training systems.\label{lastcontentpage}